%% file: main.tex
\documentclass[twoside,11pt]{article}
\input{preamble}

\ShortHeadings{Certification of Gradient-Based Training}{Sosnin, Wicker, Collyer, Tsay}
\firstpageno{1}

\begin{document}

\title{Abstract Gradient Training: A Unified Certification Framework for Data Poisoning, Unlearning, and Differential Privacy
}

\author{
\hspace{-0.6em} Philip Sosnin$^1$ \email p.sosnin23@imperial.ac.uk 
\AND \name Matthew Wicker$^1$ \email m.wicker@imperial.ac.uk
\AND \name Josh Collyer$^2$ \email jcollyer@turing.ac.uk
\AND \name Calvin Tsay$^1$ \email c.tsay@imperial.ac.uk\\
\addr $^1$Imperial College London, 
\addr $^2$The Alan Turing Institute.
}


\maketitle

\begin{abstract}
The impact of inference-time data perturbation (e.g., adversarial attacks) has been extensively studied in machine learning, leading to well-established certification techniques for adversarial robustness.
In contrast, certifying models against training data perturbations remains a relatively under-explored area.
These perturbations can arise in three critical contexts: adversarial data poisoning, where an adversary manipulates training samples to corrupt model performance; machine unlearning, which requires certifying model behavior under the removal of specific training data; 
and differential privacy, where guarantees must be given with respect to substituting individual data points.
This work introduces Abstract Gradient Training (AGT), a unified framework for certifying robustness of a given model and training procedure to training data perturbations, including bounded perturbations, the removal of data points, and the addition of new samples.
By bounding the reachable set of parameters, i.e., establishing provable parameter-space bounds, AGT provides a formal approach to analyzing the behavior of models trained via first-order optimization methods.
\end{abstract}


\subfile{sections/01_introduction}

\subfile{sections/02_related_works}

\subfile{sections/03_preliminaries}

\subfile{sections/04_problem_formulation}

\subfile{sections/05_agt}

\subfile{sections/06_propagation}

\subfile{sections/07_optimization}

\subfile{sections/08_results_basic}

\subfile{sections/09_results_poisoning}

\subfile{sections/10_results_privacy}

\subfile{sections/11_conclusions}

\acks{This work was supported by the Turing’s Defence and Security programme through a partnership with the UK government in accordance with the framework agreement between GCHQ \& The Alan Turing Institute.  CT was also supported by a BASF/Royal Academy of Engineering Senior Research Fellowship. The authors thank Fraser Kennedy and Jodie Knapp for helpful discussions and feedback, especially in the final stages of this project.}

\bibliography{references}

\appendix

\subfile{sections/12_appendix_ibp}

\subfile{sections/13_appendix_crown}

\subfile{sections/14_appendix_experimental}

\subfile{sections/15_appendix_proofs}

\end{document}

%% file: preamble.tex
\usepackage{url}            
\usepackage{booktabs}       
\usepackage{nicefrac}       
\usepackage[utf8]{inputenc} 
\usepackage[T1]{fontenc}    
\usepackage{amsfonts}       
\usepackage{amsmath}
\usepackage{amsthm}
\usepackage{amssymb}
\usepackage{nicefrac}       
\usepackage{microtype}      
\usepackage{xcolor}         
\usepackage{graphicx}
\usepackage{enumitem}
\usepackage{subcaption}
\usepackage{tabularray}
\usepackage{algorithm}
\usepackage{array} 
\usepackage{algorithmic}
\usepackage{multirow}
\usepackage{tabularx}
\usepackage{makecell}
\usepackage{bbm}
\usepackage{graphicx}
\usepackage{natbib}   
\usepackage{bbm}
\usepackage{marginnote}
\usepackage{ifthen}
\usepackage[strict]{changepage}
\usepackage{xargs}
\usepackage{marginnote}
\usepackage{wrapfig}
\usepackage[export]{adjustbox}

\usepackage{jmlr2e}
\usepackage{lastpage}
\usepackage{subfiles}

\newcommandx{\todo}[1]{%
    \checkoddpage
    \ifoddpage
      {\marginpar{\raggedright \fbox{\parbox{2.3cm}{\scriptsize \textcolor{orange}{\textbf{TODO:} #1}}}}}
    \else
      {\reversemarginpar \marginnote{\raggedright \fbox{\parbox{2.3cm}{\scriptsize \textcolor{orange}{\textbf{TODO:} #1}}}}}
    \fi
}

\DeclareMathOperator*{\argmin}{arg\,min}

\newcommand{\interval}[1]{\ensuremath{\left[#1_L, #1_U\right]}}
\newcommand{\intervalshort}[1]{\ensuremath{[#1_L, #1_U]}}

%% file: sections/01_introduction.tex
\section{Introduction}

The proliferation of machine learning models in critical applications, ranging from healthcare to autonomous driving, has raised significant concerns regarding their safety and privacy \citep{bommasani2021opportunities}.
The scale of modern datasets, while enabling unprecedented model performance, introduces significant vulnerabilities.
For example, the impracticality of exhaustive quality checks exposes systems to \textit{adversarial data poisoning} \citep{tian2022comprehensive}, while the sensitive nature of user-collected data requires rigorous guarantees on \textit{privacy leakage} \citep{dwork2014algorithmic}.
The changing landscape of data protection regulations further complicates this picture, as model owners may be required to honor individuals' \textit{right to be forgotten} \citep{bourtoule2021machine}.
These challenges present an urgent need for methods to analyze how various training-time perturbations -- including data poisoning, data removal, or data substitution -- impact model performance and security.
Understanding these effects is crucial for developing machine learning systems that remain both effective and trustworthy in sensitive applications.

The vulnerability of deep learning models to inference-time adversarial attacks is well-documented \citep{szegedy2013intriguing}, leading to significant advancements in certification and robust training methodologies \citep{konig2024critically}.
In contrast, the problem of certifying model robustness against training-time  perturbations remains comparatively under-explored.
Unlike inference-time attacks, which manipulate inputs without altering model parameters, training-time perturbations directly affect the learning process, potentially influencing the entire model architecture and all subsequent model predictions (e.g., as in Figure~\ref{fig:perturbation}).
Existing techniques, such as those for data poisoning \citep{rosenfeld2020certified, steinhardt2017certified, xie2022uncovering} or differentially private prediction \citep{papernot2016semi, van2020trade}, often rely on overly conservative analysis or require modification of the training procedure itself (e.g., requiring large ensembles of models).

This work introduces our novel framework, termed \textbf{Abstract Gradient Training (AGT)}, for efficiently analyzing the effect of data perturbation on training pipelines.
Inspired by inference-time certification techniques, AGT allows for the computation of \textit{valid bounds on the reachable set of parameters}, that is, the region of parameter space that contains all possible learned model parameters resulting from any allowable perturbation to the training data.
By shifting the focus from dataset perturbations to parameter perturbations, we show that AGT offers a more tractable path to certification.
We presented preliminary versions of this work in \cite{sosnin2024certified} and \cite{wicker2024certification}.
More recently, \cite{lorenz2024bicert,lorenz2024fullcert} apply similar methodologies to certify against data poisoning attacks. 
This work builds upon these previous works by introducing a comprehensive framework of perturbation models and certification techniques.
The contributions of this work are threefold:
\begin{itemize}
    \item We present a comprehensive framework that unifies certification for data poisoning, differentially private prediction, and certified machine unlearning, building upon and extending the foundational work of \cite{sosnin2024certified} and \cite{wicker2024certification}.
    \item We generalize the AGT framework to alternative bounding mechanisms such as mixed-integer programming (MIP), yielding tighter and more precise parameter-space bounds.
    \item We provide an extensive empirical evaluation, comparing the efficacy of various bound computation methods, including MIP-based approaches, against baselines of interval bound propagation (IBP) and linear programming (LP).
\end{itemize}
The remainder of this paper is structured as follows.
We begin with related work and mathematical preliminaries in Sections~\ref{sec:related_works} and \ref{sec:prelims}, followed by a formal problem definition in Section~\ref{sec:problem_formulation}.
We then introduce Abstract Gradient Training in Section~\ref{sec:agt}, detailing its interval bound propagation instantiation in Section~\ref{sec:ibp} and refinement via mixed-integer programming in Section~\ref{sec:optimization}.
Sections~\ref{sec:results_basic}-~\ref{sec:results_privacy} demonstrates the application of our bounds to certify properties in data poisoning, unlearning, and privacy, along with a comprehensive experimental evaluation.

%% file: sections/02_related_works.tex
\section{Background and Related Works}\label{sec:related_works}

Our methodology builds on established certification techniques developed for inference-time adversarial robustness (also known as evasion attacks), instead leveraging these methods to certify against training-time perturbations.
In addition to inference-time certification, we examine prior work on analyzing, certifying, and defending against training-time perturbations in three key areas: adversarial data poisoning, machine unlearning, and differential privacy.
This section reviews existing approaches in these domains and identifies open challenges that motivate our work.

\subsection{Certification of Adversarial Robustness}\label{sec:related_works_verif}

\begin{figure}[t]
    \centering
    \includegraphics[trim={1cm 0 1cm 0}, width=\linewidth]{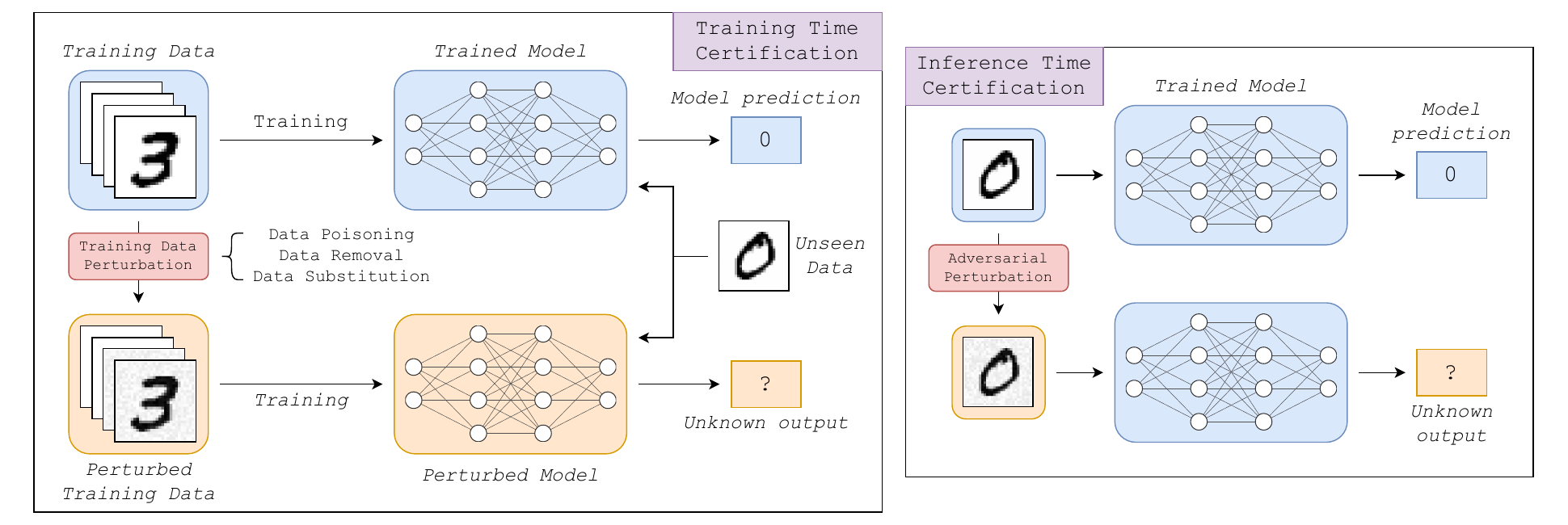}
    \caption{Illustration of training-time certification (left) vs inference-time certification (right). The goal of training-time certification is to verify the behavior of the perturbed model for any possible training data perturbation. Unlike inference-time certification, training-time certification requires reasoning over the entire training process.}
    \label{fig:perturbation}
\end{figure}

Machine learning models, particularly neural networks, are susceptible to adversarial examples -- carefully designed inputs intended to deceive the model, often by introducing changes imperceptible to humans \citep{szegedy2013intriguing}.
Even well-trained networks that achieve high test-time accuracy may fail to generalize to new scenarios, leaving them vulnerable to adversarial attacks \citep{papernot2016limitations}.
In applications of machine learning for which incorrect predictions can lead to significant consequences, it is crucial to develop provable verification techniques to complement empirical performance testing.

Formally, verification aims to determine whether a machine learning model adheres to a specified input-output relationship over a given input domain \citep{liu_algorithms_2019}.
Methods can be categorized based on the strength of their guarantees:
\begin{itemize}
    \item \textit{Sound} algorithms are those that will only declare a property to be true if it is true.
    \item \textit{Complete} algorithms are guaranteed to confirm a property is true whenever it is true.
\end{itemize}
While soundness and completeness are both desirable, to improve computational efficiency, some approaches sacrifice completeness by employing approximations, potentially leading to \textit{incomplete} results. In this work, we will use the following terminology: \textit{Verification} will refer to strictly methods that are both sound and complete, and \textit{certification} will describe methods that are sound but incomplete. 

\paragraph{Existing Approaches.}
Certification techniques can be broadly categorized based on the techniques they use. Those based on propositional logic encode the network and property as a satisfiability problem. If the formula is satisfiable, it implies that a counterexample exists, meaning the property is violated \citep{katz2017reluplex}. Alternatively, those that use domain propagation or abstract interpretation propagate reachable domains through the network's layers. This technique over-approximates the set of possible outputs for a given input domain. If the propagated output domain satisfies the desired property, it therefore holds over the input domain \citep{gowal2018effectiveness, xiang2018output, zhang2018efficient, singh2019}.

Finally, optimization-based methods encode neural networks as a set of constraints, transforming the certification task into an optimization problem \citep{ehlers_formal_2017, fischetti2018deep, bunel2018unified}.
These techniques commonly leverage mixed-integer programming solvers to either certify robustness or identify counterexamples. We refer the reader to \cite{huchette2023deep} for a review of these encodings and problem formulations.

\paragraph{Relation to This Work.} Existing certification algorithms are not directly applicable to certifying models with respect to training-time perturbations.
Our certification problem specifically involves analyzing the entire training trajectory while accounting for perturbations at each iteration.
Viewing the training process as a non-linear dynamical system, our approach aligns with existing literature on reachability analysis for such systems \citep{althoff2021set}

In our case, the reachability analysis must encompass not only the neural network itself, but also the computation of gradients and parameter updates at each training step.
The most closely related certification approaches are those that treat both the model’s inputs and parameters as variables.
Such formulations appear in the certification of Bayesian neural networks \citep{wicker2020probabilistic} and in preliminary work on robust explanations \citep{wicker2022robust}.
However, despite these methodological connections, none of these existing approaches directly extend to the general training setting considered in this work.

\subsection{Adversarial Data Poisoning}

Data poisoning attacks represent a significant threat to the integrity of machine learning models, wherein malicious actors manipulate training data to degrade model performance or induce specific, undesirable behaviors \citep{biggio2018wild, biggio2014poisoning, newsome2006paragraph}.
A backdoor attack, for example, introduces a hidden vulnerability where a specific input pattern, or trigger, forces the model to act in an unexpected way at inference time, while the model behaves normally on all other inputs.
For example, a lane detection system with a backdoor might be tricked into misclassifying lane markings if it sees a traffic cone \citep{han2022physical}.

Previous research has revealed that attacks affecting even a small proportion of training data can lead to catastrophic model failures \citep{carlini2023poisoning}. 
For example, poisoning can readily manipulate recommender systems on platforms such as YouTube, eBay, and Yelp \citep{yang2017fake}.
Other research has shown that poisoning just 1\% of training data can force targeted misclassifications \citep{zhu2019transferable}, or introduce targeted backdoor vulnerabilities in lane detection systems \citep{han2022physical}.

Poisoning attacks can be broadly categorized based on their objectives:
\textit{untargeted poisoning} aims to generally corrupt model performance, potentially leading to denial-of-service \citep{munoz2017towards}. \textit{Targeted poisoning} focuses on misclassifying specific inputs while maintaining overall performance, and \textit{backdoor attacks} introduce hidden triggers that activate only when particular patterns are present at inference time \citep{chen2017targeted, gu2017badnets, han2022physical, zhu2019transferable}. In this work, we adopt the classification of attack goals proposed in \cite{tian2022comprehensive}, and refer readers to that source for a more detailed review of data poisoning attacks.

\paragraph{Adversary Capabilities.}
Modeling the capabilities of a poisoning adversary is critical in assessing the safety of machine learning systems.
In this work, we consider two distinct threat models: bounded and unbounded attacks.
In \textit{bounded attacks}, adversaries are constrained in the magnitude of perturbations they can apply to training data, typically defined by norms in feature and label spaces.
In contrast, \textit{unbounded attacks} allow adversaries to inject arbitrary data points, potentially exploiting data collection vulnerabilities.
The latter assumes a more powerful adversary capable of significantly altering the training distribution.
In both cases, we assume a white-box setting where the adversary possesses comprehensive knowledge of the training process, including model architecture, initialization, data, and hyperparameters, to establish worst-case robustness guarantees.

\paragraph{Poisoning Attack Defenses.}
Defending against poisoning attacks is a complex challenge. Traditional defenses often focus on identifying and mitigating specific attack strategies.
For example, \cite{li2020learning} train classifiers to detect and reject potentially poisoned inputs by leveraging datasets generated from known attack strategies.
Other approaches apply noise or clipping techniques to limit the impact of certain perturbations \citep{hong2020effectiveness}. However, these targeted defenses may not generalize to novel or more sophisticated attacks.

To address the limitations of attack-specific defenses, researchers have explored methods for \textit{certified robustness}, which aim to provide provable guarantees against a wider range of poisoning attacks.
For linear models, \cite{steinhardt2017certified} and \cite{rosenfeld2020certified} establish upper bounds on the effectiveness of gradient-based and $\ell_2$ perturbation attacks, respectively.
Differential privacy has also been investigated as a means to provide statistical guarantees in limited poisoning scenarios \citep{xie2022uncovering}.
Deterministic certification methods, such as those employing \textit{aggregation} techniques, involve partitioning the dataset and training multiple models to achieve robustness guarantees \citep{levine2020deep, wang2022improved, rezaei2023run}.
These methods offer strong guarantees but often incur substantial computational costs given the need to train and evaluate numerous models.

Most similar to this work is the recent work of \cite{lorenz2024fullcert} and \cite{lorenz2024bicert}. In these works, they apply interval bound propagation and mixed-integer programming to certify training pipelines against data poisoning attacks.
While methodologically similar, these works are more limited in scope and thus can be considered as a subset of the more general approach we present here.
Their work focuses on a reduced set of threat models (specifically, bounded perturbations applied to all training samples), while we consider a wider range of adversaries, including those limited to poisoning a small number of samples with unbounded perturbations. Furthermore, their analysis is confined to a single training iteration using a mixed-integer bilinear formulation, whereas our work examines multiple training iterations and a wide range of formulations and relaxations, including linear and quadratic approximations. Finally, we also extend the certification framework to a wider family of parameter domains, such as polyhedral domains.

\paragraph{Relation to This Work.} Unlike many prior works, our approach aims to certify a given model and training algorithm and therefore can be used to analyze and certify the poisoning robustness of various proposed defenses and benchmarks.
Like existing approaches, our method computes a sound (but potentially incomplete) certificate that bounds the impact of the poisoning attack.
Importantly, our work is complementary to prior work on aggregation methods, suggesting potential avenues for future research to combine both techniques for enhanced, certified robustness.

\subsection{Differential Privacy}

The deployment of machine learning in sensitive domains, such as healthcare and finance, requires robust privacy safeguards.
As these models increasingly rely on personal data, ensuring the confidentiality of individuals' information becomes paramount \citep{cummings2021need}.
Differential privacy (DP) has emerged as the primary tool in the development of privacy-preserving machine learning algorithms.
DP provides a rigorous mathematical framework to quantify and limit the leakage of sensitive information from datasets.
Formally, differential privacy is defined in terms of the probability of obtaining a specific set of outcomes from a randomized algorithm when applied to adjacent datasets.
\begin{definition}[$(\epsilon, \delta)$-Differential Privacy \citep{dwork2014algorithmic}]\label{def:approxdp} 
    A randomized mechanism $\mathcal{M}$ is $(\epsilon, \delta)$-differentially private if, for all pairs of adjacent datasets $A, B$ and any $S \subseteq \operatorname{Range}(\mathcal{M})$, 
    \begin{equation}
        \mathbb{P}\big(\mathcal{M}(A) \in S \big) \leq e^\epsilon\mathbb{P}\big(\mathcal{M}(B) \in S \big) + \delta
    \end{equation}
Here, $\epsilon$ is known as the privacy leakage, while $\delta$ allows for a small probability of failure.
\end{definition}
In this work, we define distances between datasets using the Hamming distance $d(A, B)$, and adjacent datasets are those for which $d(A, B) = 1$.

\paragraph{Differential Privacy in Machine Learning.}
Current approaches to achieve differential privacy in machine learning broadly fall into two categories: private training and private prediction.
Private training methods, exemplified by DP-SGD \citep{abadi2016deep}, introduce randomness into the training process to ensure that it satisfies Definition~\ref{def:approxdp}, thereby enabling the release of model parameters with differential privacy guarantees.
This approach, while effective, incurs costs due to the required pre-specification of privacy parameters, and it risks introducing harmful biases during training, such as discriminatory effects \citep{fioretto2022differential}.

Private prediction, on the other hand, focuses on privatizing the model's output predictions.
It allows for dynamic adjustment of privacy budgets and is applicable to complex training scenarios, such as federated learning.
Existing techniques for private prediction, e.g. the subsample-and-aggregate approach \citep{papernot2016semi, van2020trade}, add noise to the model's predictions calibrated to the \textit{global sensitivity}.
This approach, while providing a straightforward guarantee of Definition~\ref{def:approxdp}, often leads to the addition of excessive noise, as global sensitivity considers the worst case across all possible datasets.
Consequently, the utility of the model's predictions can be significantly and unnecessarily diminished, particularly when the model's behavior is locally stable.

In practice, private prediction has been observed to exhibit an empirically unfavorable privacy-utility trade-off compared to private training \citep{van2020trade}, and recent audits have indicated a lack of tightness in their privacy analyses \citep{chadha2024auditing}.

\paragraph{Limitations of Current Approaches to Differential Privacy.}
Despite DP's widespread adoption, \textit{post-hoc} audits have revealed gaps between attacker strength and guarantees offered by DP \citep{carlini2022membership, yu2022individual}.
As a result, several works seek more specific, and thus sharper, privacy guarantees by leveraging specific information about the learning algorithm and dataset.
For example, \citet{nissim2007smooth} and \citet{liu2022differential} use notions of local sensitivity to produce tighter bounds.
In \cite{ligett2017accuracy}, the authors privately search the space of privacy-preserving parameters to tune performance on a given dataset, while in \cite{yu2022individual} the authors propose individual differential privacy, which can compute tighter privacy bounds for given individuals in the dataset.
Unlike this work, the above works rely solely on private training, e.g., DP-SGD \citep{abadi2016deep}.

\paragraph{Relation to This Work.}
An application of the methodology proposed in this work is proving tighter bounds for the private prediction setting.
Specifically, we aim to use certification to prove the stability of given predictions of machine learning models, thus providing data-specific bounds on the (local) sensitivity of model predictions.
We then build upon the approach of \cite{nissim2007smooth} to derive bounds on the \textit{smooth sensitivity} of model predictions, which we use to improve the privacy-utility tradeoffs inherent in the private prediction setting.
This represents a significant step towards bridging the performance gap between private prediction and private training.

\subsection{Machine Unlearning.}

Related to differential privacy is the concept of machine unlearning, which addresses complementary privacy concerns.
While DP focuses on preventing any information leakage from a model's training data, machine unlearning is concerned with only the removal of specific data points' influence \citep{cao2015towards}.
This capability has become increasingly important due to regulations such as GDPR in the EU, which codify the ``right to be forgotten'' \citep{bourtoule2021machine}.
As models grow in size and complexity, naive retraining approaches become prohibitively expensive, requiring more efficient unlearning methods.

Unlearning guarantees can be broadly categorized into two types.
\textit{Exact unlearning} \citep{bourtoule2021machine} requires that the unlearned model be identical to a model retrained from scratch without the forgotten data.
Since this is often computationally infeasible, \textit{approximate unlearning} \citep{guo2019certified} relaxes this requirement by instead limiting statistical similarity between the unlearned model and the retrained one, formalized using bounds similar to differential privacy. 
In fact, differentially private learning algorithms inherently satisfy approximate unlearning guarantees, though often with excessive performance penalties \citep{neel2021descent}.
Within this category, certified unlearning \citep{guo2019certified} aims to provide verifiable certificates that bound the statistical distance between the unlearned and retrained models, i.e. to provide guarantees in the context of approximate unlearning.
These approaches provide provable bounds on the statistical distance between unlearned models and those trained from scratch without the forgotten data point.

\paragraph{Existing Approaches.}
Current certified unlearning techniques include influence function-based methods \citep{koh2017understanding, guo2019certified}, which approximate the effect of removing training points through first-order Taylor expansions of the model's loss function.
While computationally efficient, these methods can produce inaccurate certificates for non-convex models \citep{basu2020influence}.
Another prominent approach is SISA (Sharded, Isolated, Sliced, and Aggregated) training \citep{bourtoule2021machine}, which partitions the training data across multiple shards and only retrains affected shards when unlearning requests arise.
This reduces computational overhead of retraining but may compromise model performance due to data fragmentation.

Despite progress in the field, existing unlearning methods face significant challenges.
Most critically, current certification methods focus on bounding the distance between model parameters, which does not directly translate to guarantees on model predictions \citep{thudi2022necessity}.
This parameter-space focus can lead to overly conservative certificates that do not accurately reflect the actual privacy risks posed by unlearned models.

\paragraph{Relation to This Work.}
While our work does not fit within either the exact or approximate unlearning settings, our approach extends the analysis of the influence of unlearning on the predictions of a machine learning model.
Unlike approaches based on influence functions, our parameter-space bounds give formal (non-approximate) guarantees on the impacts of data removal.
Furthermore, rather than working solely in parameter space, we use our bounds to certify aspects of model behavior, such as how predictions change when data is removed.
This allows for a precise characterization of model behavior under unlearning.

%% file: sections/03_preliminaries.tex
\section{Preliminaries}\label{sec:SGD}\label{sec:prelims}

Before introducing our certification framework, we first establish the necessary mathematical background and notation.

\paragraph{Notation.}
We define a machine learning model as a function $f(x, \theta)$ with parameters $\theta$ mapping inputs $x$ from a feature space to outputs $y$ in a target space.
We consider both regression and classification settings, allowing the target space to be discrete, continuous, and/or multivariate.
Our focus is on supervised learning, where the model is trained on a labeled dataset $\mathcal{D} = \{ (x^{(i)}, y^{(i)}) \}_{i=1}^{N}$ consisting of $N$ input-output pairs used to optimize the parameters $\theta$.

\paragraph{Neural networks.}
While the methods introduced in this work are applicable to a broad class of machine learning models, we primarily focus on feed-forward neural networks.
A feed-forward neural network if a machine learning model $f$ defined as a composition of $K$ layers with parameters $\theta=\left\{(W^{(i)}, b^{(i)})\right\}_{i=1}^K$ given by
\begin{align}
    \hat{z}^{(k)} = W^{(k)} z^{(k-1)} + b^{(k)}, \quad z^{(k)} = \sigma \left(\hat{z}^{(k)}\right),
\end{align}
where $z^{(0)} = x$, $f(x, \theta) = \hat{z}^{(K)}$, and $\sigma$ is the activation function, which we take to be the rectified linear unit (ReLU).

\paragraph{Gradient-based training}
Training neural networks typically relies on \textit{gradient-based optimization}, where model parameters are iteratively updated to minimize a loss function.
Let $\mathcal{L}$ denote a loss function that measures model quality, e.g. how different the model predictions are from the ground truth labels.
Given a training dataset $\mathcal{D}$, the model parameters $\theta$ are optimized by minimizing the empirical risk:
\begin{equation}
\theta^* = \argmin_{\theta} \frac{1}{N} \sum_{i=1}^{N} \mathcal{L}\left(f\left(x^{(i)}, {\theta}\right), y^{(i)}\right).
\end{equation}
The predominant approach to solving this optimization problem is through \textbf{stochastic gradient descent (SGD)}, where the model parameters are iteratively updated using a randomly sampled mini-batch $\mathcal{B} \subset \mathcal{D}$. The update rule is given by:
\begin{equation}
\theta^{(t)} = \theta^{(t-1)} - \frac{\alpha}{|\mathcal{B}|} \sum_{(x^{(i)}, y^{(i)}) \in \mathcal{B}} \nabla_\theta \mathcal{L}\left(f\left(x^{(i)}, {\theta^{(t-1)}}\right), y^{(i)}\right),\label{eq:sgd}
\end{equation}
where $\alpha > 0$ is the learning rate, and the gradient is averaged over the mini-batch $\mathcal{B}$.
We will denote the parameters obtained by following this procedure for a model $f$ with a dataset $\mathcal{D}$ as $\theta = M\left(\mathcal{D}\right)$.

Several extensions of SGD introduce adaptive learning mechanisms to enhance convergence, such as Adam~\citep{kingma2014adam}.  
While the techniques developed here apply to any first-order optimization algorithm, we focus on standard SGD \eqref{eq:sgd} to simplify our exposition.

%% file: sections/04_problem_formulation.tex
\section{Problem Formulation}\label{sec:problem_formulation}

In this section, we formalize the problem of certifying robustness to training data perturbations and define perturbation models relevant to the applications of data poisoning, machine unlearning and differential privacy.
These formulations serve as the basis for our certification framework, which we describe in the subsequent sections.

\subsection{Certification Problem}
In the context of training data perturbations, our certification objective is to formally certify whether a machine learning model trained with a specified algorithm satisfies a given property despite modifications to its training data.
Given an initial dataset $\mathcal{D}$, a training procedure $M$, and a perturbation model $\mathcal{T}$ that defines feasible training data modifications, we seek to certify that a property $P$ holds for any trained parameter $\tilde{\theta}$ obtained from a perturbed dataset $\tilde{\mathcal{D}} \in \mathcal{T}(\mathcal{D})$.
Formally, we aim to prove:
\begin{equation}
\forall \tilde{\mathcal{D}} \in \mathcal{T}(\mathcal{D}), \quad \tilde{\theta} = M(\tilde{\mathcal{D}}) \implies P(\tilde{\theta}).
\end{equation}
If this condition holds, the model is guaranteed to satisfy property $P$ across all valid training data perturbations, establishing a formal guarantee under the specified perturbation model.

Following standard approaches in the machine learning robustness literature (see Section~\ref{sec:related_works_verif}), we reformulate this certification problem as an optimization problem:
\begin{equation}
\max_{\tilde{\mathcal{D}}} \ J(\tilde{\theta}) \quad \text{s.t.} \quad \tilde{\theta} = M(\tilde{\mathcal{D}}), \quad \tilde{\mathcal{D}} \in \mathcal{T}(\mathcal{D}).\label{eq:cert_problem}
\end{equation}
where $J(\theta)$ is an objective function related to the property $P$ via a specification on the optimal solution of the above problem. For instance, if $P$ is the correctness of the model's prediction on a particular input $x$, one can reformulate the certification problem via $J(\theta)=c^Tf(x, \theta)$ for some vector $c$, i.e. a linear specification on the output logits of the network \citep{gowal2018effectiveness}.
If the optimal value of the optimization problem is $\leq b$ for a chosen constant $b$ then we can say that $P$ is guaranteed to hold.

In this work, we focus on \textit{sound but incomplete} certification, meaning that we aim compute a provably valid upper bound on the above optimization problem.
If this bound remains within some specification determined by $P$, we establish a formal certification guarantee.

In the following subsection, we outline specific choices for $\mathcal{T}$ that correspond to key applications in data poisoning robustness, machine unlearning and differential privacy.
Choices of objective function $J$, and the practical implications of post-training certificates are discussed in detail in Sections~\ref{sec:results_basic}-~\ref{sec:results_privacy}.

\subsection{Characterizing Training Data Perturbations}

A \textit{perturbation model} $\mathcal{T}(\mathcal{D})$ specifies the set of all possible modified datasets that could result, e.g., from adversarial manipulation or privacy-driven modifications.
We categorize training data perturbations into three primary types:

\paragraph{1) Bounded manipulation of training data.} In the bounded perturbation setting, we assume that any subset of up to $n$ training samples may be modified within a predefined constraint, such as an $\ell_p$-norm ball.
Given a nominal dataset $\mathcal{D}$, we define the perturbed dataset as $\tilde{\mathcal{D}} \in \mathcal{T}^{n, p, \epsilon, q, \nu}_{\text{bounded}}(\mathcal{D})$, where perturbations are constrained by:  
\begin{equation}
\| x^{(i)} - \tilde{x}^{(i)} \|_p \leq \epsilon, \quad \| y^{(i)} - \tilde{y}^{(i)} \|_q \leq \nu, \quad \forall i \in \mathcal{I}.\label{eq:bound_adv}
\end{equation}
Here, $\epsilon$ and $\nu$ represent the maximum allowable perturbations in the feature and label spaces, respectively, measured using the $\ell_p$ and $\ell_q$ norms.
The index set $\mathcal{I}$ specifies the subset of up to $n$ training samples that can be modified, which we take to be any set satisfying $\mathcal{I} \subset \{1, \dots, N\}$ subject to $|\mathcal{I}| \leq n$.
This perturbation model encompasses any poisoning adversary capable of modifying up to $n$ training data samples within the specified bounds.

\paragraph{2) Removal of training data.}
In training data removal, the perturbation model considers datasets $\tilde{\mathcal{D}} \in \mathcal{T}^n_{\text{removal}}(\mathcal{D})$, where up to $n$ samples are removed from the original dataset:
\begin{equation}
\tilde{\mathcal{D}} = \mathcal{D} \setminus \mathcal{S}, \quad |\mathcal{S}| \leq n,
\end{equation}
where $\mathcal{S} \subset \mathcal{D}$ is the set of removed data-points.
This setting is particularly relevant for machine unlearning applications, where certification guarantees must be given with respect to the deletion of specific training data points.

\paragraph{3) Replacement of training data.}
The substitution model generalizes both the bounded poisoning and removal settings by allowing an adversary to replace a subset of the training data with new points, leading to a dataset $\tilde{\mathcal{D}} \in \mathcal{T}^n_{\text{subs.}}(\mathcal{D})$ of the form:
\begin{equation}
\tilde{\mathcal{D}} = (\mathcal{D} \setminus \mathcal{S}) \cup \tilde{\mathcal{S}}, \quad |\mathcal{S}| = |\tilde{\mathcal{S}}| \leq n,
\end{equation}
where $\mathcal{S}$ is a set of up to $n$ removed samples and $\tilde{\mathcal{S}}$ is a set of newly introduced samples.
This type of perturbation allows consideration of both general poisoning attacks (where adversaries inject maliciously crafted data that is not necessarily related to the removed samples) or differential privacy (where guarantees must be given with respect to the substitution of data points).

%% file: sections/05_agt.tex
\section{A Unified Certification Framework}\label{sec:agt}

\begin{figure}
    \centering
    \includegraphics[width=\linewidth]{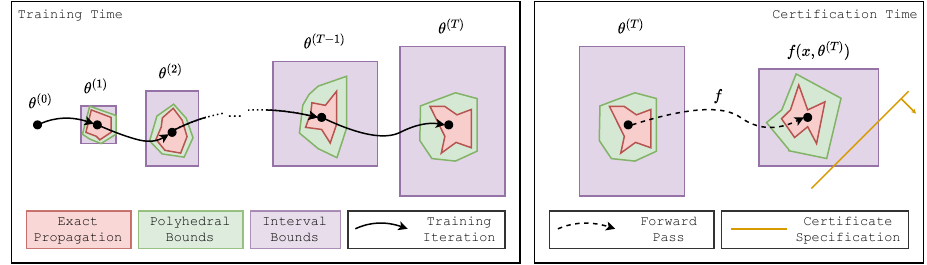}
    \caption{
    Outline of our certification framework.
    During training, parameter-space bounds are propagated through each iteration, admitting some over-approximation.
    Post-training, the final parameter-space bounds are used to certify the network against a given specification. While we illustrate certification in logit space, our framework supports certification with respect to any desired criterion.}
\end{figure}

Analyzing the behavior of machine learning models under training data perturbations poses a significant theoretical and computational challenge. 
Unlike standard inference-time certification, which considers perturbation to a single data-point, our setting considers the worst-case outcome in the space of all possible datasets that could have been used to train the model.
This significantly expands the complexity of the analysis, as it requires reasoning over a combinatorially large space of possible training datasets, each of which can lead to a different set of learned model parameters.
To overcome this challenge, we introduce the concept of valid parameter-space domains, which provide a structured way to efficiently certify properties of the trained model.
We then present Abstract Gradient Training (AGT), a framework designed to compute these bounds and enable scalable certification.

\subsection{Valid Parameter-Space Domains}  

Certification problems of the form \eqref{eq:cert_problem} present a significant computational challenge, as they require optimization over the space of all possible training datasets.  
Rather than directly optimizing over this combinatorially large space, we instead seek to establish a \textit{valid parameter-space domain}, which defines the reachable set in parameter space that contains all possible learned model parameters resulting from any training with any allowable perturbation to the training data.
This reformulation allows us to shift our certification problem from dataset perturbations to parameter perturbations, enabling more efficient analysis. 

Formally, let $\mathcal{D}$ be a nominal training dataset, and let $\mathcal{T}(\mathcal{D})$ denote the set of all reachable datasets under a given perturbation model (i.e., modification, removal, or substitution).  
We define a set $\Theta$ to be a \textbf{valid parameter-space domain} with respect to a training algorithm $M$ if:
\begin{equation}
\tilde{\theta} = M(\tilde{\mathcal{D}}) \in \Theta, \quad \forall \tilde{\mathcal{D}} \in \mathcal{T}\big(\mathcal{D}\big),\label{eq:parambounds}
\end{equation}
where $\tilde{\theta} = M(\tilde{\mathcal{D}})$ represents the parameters obtained by training on a perturbed dataset $\tilde{\mathcal{D}}$.
Thus, $\Theta$ serves as a certified enclosure that bounds all possible learned parameters resulting from any feasible modification to the training dataset.

Any parameter-space domain satisfying \eqref{eq:parambounds} enables us to upper-bound our certification objective in \eqref{eq:cert_problem}.
Specifically, we establish the following key result:

\begin{theorem}\label{thm:paramcerts}
Let $\Theta$ be a valid parameter-space domain for a given perturbation model $\mathcal{T}$.  
Then, for any objective function $J$, the worst-case impact of training data perturbations can be bounded by optimizing over the parameter space instead of the dataset space:
\begin{equation}\label{eq:translationthm}
    \max\limits_{\tilde{\mathcal{D}} \in \mathcal{T}(\mathcal{D})} J\big(M\big(\mathcal{\tilde{D}}\big)\big) \ \leq \ \max\limits_{\tilde{\theta} \in \Theta}  J\big( \tilde{\theta}\big).
\end{equation}
\end{theorem}

Theorem~\ref{thm:paramcerts} establishes that we can shift certification problems over the adversarial dataset perturbation into the parameter space, where it can be more tractably analyzed.
Unlike the intractable dataset-space maximization on the left-hand side of \eqref{eq:translationthm}, the equivalent optimization problem over the parameter-space domain on the right-hand side can be efficiently upper-bounded using established certification techniques \citep{adams2023bnn, wicker2020probabilistic, wicker2023adversarial}.  
This transformation constitutes the basis for our approach to certified robustness and privacy guarantees.

\subsection{Abstract Gradient Training for Parameter-Space Bounds}

In this section, we outline our framework for computing valid parameter-space domains.
Rather than attempting to track the exact optimization trajectories of model parameters under possible training data perturbations, we construct an \textbf{abstract representation} of the training dynamics that provides a sound over-approximation of all reachable parameter values, i.e., bounds on the reachable set.
This approach, which we term \textbf{Abstract Gradient Training (AGT)}, effectively allows us to compute provable bounds on the set of parameters that a model can reach under a given perturbation model.

To begin, we impose the following assumptions on our analysis\footnote{While it is possible to relaxed these assumptions within our framework, this leads to increased computational cost and reduced certification tightness.}.:
\begin{itemize}
    \item \textbf{Assumption 1: Fixed data ordering.} The sequence in which training samples are processed remains unchanged between the nominal and perturbed datasets
    \item \textbf{Assumption 2: Fixed parameter initialization.} The training process starts from the same parameter initialization for both the nominal and perturbed datasets.
    \item \textbf{Assumption 3: Fixed training hyperparameters.} The perturbation model cannot influence hyperparameters such as learning rate, batch size, or number of training iterations.
\end{itemize}  
These assumptions allow us to isolate the effect of training data perturbations while eliminating variability due to factors such as stochastic parameter initialization or data shuffling.  
As a result, our certification framework provides guarantees that specifically account for adversarial modifications to the training data.
It is important to note that any certificates derived under this framework are valid only within the scope of these assumptions.

\paragraph{Abstract Training Dynamics.}  
Let $\theta^{(t)}$ denote the model parameters at training step $t$ under standard gradient-based optimization.  
Given a loss function $\mathcal{L}$, the parameters are updated using the stochastic gradient descent (SGD) rule, reproduced from \eqref{eq:sgd}:
\begin{equation}
\theta^{(t)} = \theta^{(t-1)} - \frac{\alpha}{|\mathcal{B}|} \sum_{(x^{(i)}, y^{(i)}) \in \mathcal{B}} \nabla_\theta \mathcal{L}\left(f\left(x^{(i)}, \theta^{(t-1)} \right), y^{(i)}, \right),
\end{equation}
where $\alpha$ is the learning rate, and $\nabla_\theta \mathcal{L}$ is gradient of the loss function with respect to the model parameters.  
When the training dataset is perturbed, the loss landscape and resulting gradient updates also shift, causing deviations in the optimization trajectory at both the \textit{current iteration} and \textit{all subsequent iterations}.  

To obtain sound bounds on the parameters under such perturbations, we define a valid parameter-space domain $\Theta^{(t)}$ at each training step $t$ such that:
\begin{equation}
\theta^{(t)}(\tilde{\mathcal{D}}) \in \Theta^{(t)}, \quad \forall \tilde{\mathcal{D}} \in \mathcal{T}(\mathcal{D}),
\end{equation}
where we use $\theta^{(t)}(\tilde{\mathcal{D}})$ to denote the value of the $t$-th parameter iterate trained using the dataset $\tilde{\mathcal{D}}$.
That is, at each step of training, the set $\Theta^{(t)}$ provides an enclosure containing all possible parameter values that could be reached when training on any perturbed dataset $\tilde{\mathcal{D}}$.
By propagating these bounds through the optimization process, we obtain a final parameter-space domain $\Theta^{(T)}$ that satisfies \eqref{eq:parambounds}.

A key challenge in AGT is propagating the parameter-space domains through successive training iterations.
Each parameter update must account for both: (1) the cumulative effect of perturbations from all previous iterations and (2) the direct effect of perturbations in the current iteration.
Formally, given $\Theta^{(t-1)}$ (the parameter-space domain at step $t-1$), the domain at step $t$ is defined as:  
\begin{equation}
    \Theta^{(t)} = \left\{\theta^{(t-1)} - \frac{\alpha}{|\tilde{\mathcal{B}}|} \sum_{\tilde{\mathcal{B}}} \nabla_\theta \mathcal{L}\left(f\left(x^{(i)}, \theta^{(t-1)}\right), y^{(i)} \right) \mid \tilde{\mathcal{B}} \subset \mathcal{T}(\mathcal{D}),\ \theta^{(t-1)} \in \Theta^{(t-1)} \right\}.
\end{equation}
Here, the constraint $\theta^{(t-1)} \in \Theta^{(t-1)}$ ensures that the domain captures the effect of training data perturbations across all previous iterations, while $\tilde{\mathcal{B}} \subset \mathcal{T}(\mathcal{D})$ represents perturbations in the current mini-batch.
The former is necessary to represent the worst case $\theta^{(t-1)} \in \Theta^{(t-1)}$ in terms of the latter step. 
In general, the exact parameter-space bound $\Theta^{(t)}$ has a complicated form, as it encapsulates the entire history of training dynamics, including past gradients, dataset perturbations, and optimization steps.
Directly maintaining and propagating this exact set is computationally intractable for all but the simplest models.  

To address this, we restrict our analysis to specific classes of \textbf{abstract domains}, which over-approximate $\Theta^{(t)}$ in a practical manner while maintaining soundness.
These abstract domains balance the trade-off between \textit{tightness}, which ensures meaningful certification guarantees, and \textit{tractability}, which enables efficient computation.
This trade-off has been thoroughly investigated in the literature on certifying adversarial robustness, for example in \cite{huchette2023deep}.
In particular, we will explore the following abstract domains:
\begin{itemize}
    \item \textbf{Intervals:} A simple yet computationally efficient abstraction, where each parameter $\theta_j^{(t)}$ is independently bounded within an interval $[\underline{\theta}_j^{(t)}, \overline{\theta}_j^{(t)}]$. This approach enables fast propagation of parameter-space bounds (i.e., a box domain) through successive training iterations, but may introduce significant over-approximation error.
    \item \textbf{Polytopes:} A more expressive abstraction that captures dependencies among parameters by over-approximating $\Theta^{(t)}$ with a convex polytope $\{\theta \mid A \theta \leq b\}$. This domain can capture dependencies between parameters but requires solving a set of convex optimization problems at each step.
    \item \textbf{Mixed-Integer Domains:} The most precise domain, where $\Theta^{(t)}$ is encoded as a set of linear, quadratic, and integer constraints. This formulation can provide tight parameter-space bounds, but is computationally expensive, requiring the solution of NP-Hard non-convex optimization problems.
\end{itemize}

Each of these domains provides a different balance between computational efficiency and certification tightness, which we explore in subsequent sections.

\subsection{Abstract Gradient Training as Constrained Optimization}

At each iteration of AGT, the exact set of reachable parameters can be expressed as the feasible region of a constrained optimization problem.  
Given an initial parameter set $\Theta^{(0)}$, we characterize the set of all possible parameters at iteration $T$ as:
\begin{alignat*}{3}
         & \theta^{(0)} \in \Theta^{(0)} \tag{Initial Conditions}\\
         &\mathcal{\widetilde{D}} = \left\{ \left(\tilde{x}^{(i)}, \tilde{y}^{(i)}\right)\right\}_{i=1}^N\in \mathcal{T}(\mathcal{D}) & \tag{Dataset Perturbation}\\
         & \mathcal{B}^{(t)} = \left\{ \left(\tilde{x}^{(i)}, \tilde{y}^{(i)}\right)\right\}_{i\in \mathcal{I}^{(t)}} \subset \mathcal{\widetilde{D}} & t\in [T] \tag{Batch Sampling} \\ 
         & \hat{y}^{(t, i)} = f\left(\tilde{x}^{(i)}, \theta^{(t-1)}\right) & i\in \mathcal{I}^{(t)}, t \in [T] \tag{Forward Pass}\\
         & \delta^{(t, i)} = \nabla_\theta\mathcal{L}\left(\hat{y}^{(t, i)}, \tilde{y}^{(i)}\right) &  i\in \mathcal{I}^{(t)}, t \in [T] \tag{Backward Pass}\\
         & \theta^{(t)} = \theta^{(t-1)} - \frac{\alpha}{|\mathcal{B}^{(t)}|}\sum\limits_{i\in \mathcal{I}^{(t)}} \delta^{(t, i)} & t\in [T] \tag{Parameter Update}
\end{alignat*}
Here, the fixed index sets $\mathcal{I}^{(t)} \subset [N]$ capture the data ordering assumption by assigning training samples to each batch $\mathcal{B}^{(t)}$.

The constraint corresponding to the forward pass has been extensively studied in the optimization literature, particularly through mixed-integer encodings of neural networks and their activation functions. In contrast, the backward pass has received significantly less attention in the context of constrained optimization. We provide exact formulations for both components of the optimization problem in Section~\ref{sec:optimization}.

The formulation above captures all possible parameter values reachable at time step $T$ under the given training data perturbations.  
As mentioned above, directly computing the exact reachable set $\Theta^{(t)}$ is intractable for large-scale models, as it requires reasoning over all possible perturbations and optimization trajectories.
Instead, this optimization-based formulation serves as a foundation from which \textit{tractable relaxations} can be derived, which will be the subject of the following sections.

%% file: sections/06_propagation.tex
\section{Efficient Relaxation via Interval Bound Propagation}\label{sec:ibp}

Interval arithmetic is a well-established technique that favors computational efficiency over tightness in the certification of machine learning models \citep{gowal2018effectiveness}. 
Interval domains provide a simple yet scalable method for over-approximating the space of possible model parameters throughout training.  
By bounding each parameter independently within an interval, this approach allows for sound propagation of bounds while maintaining computational tractability.

\subsection{The Interval Domain}

In an interval-based representation, each parameter $\theta_j$ is independently bounded within an interval.
For ease of notation, we denote a full set of parameter intervals as the interval domain $\boldsymbol{\theta} = \left[{\theta^L}, {\theta^U}\right]$, where membership is defined as:
\begin{equation}
\theta \in \boldsymbol{\theta} \implies {\theta}_j^L \leq \theta_j \leq {\theta}_j^U, \quad \forall j.
\end{equation}
Interval domains offer a computationally efficient alternative to more complex abstractions, such as polytopes or mixed-integer domains.
However, they can introduce significant \textit{over-approximation error} due to the inherent independence assumption between parameters.
Despite this limitation, interval bounds serve as a useful baseline for certifying robustness, particularly in large-scale models where finer-grained representations may be intractable.

\paragraph{Interval Arithmetic.}
To efficiently propagate interval-based bounds, we leverage the well-established framework of \textbf{interval arithmetic} \cite{rump1999fast}, which extends standard arithmetic operations to interval-valued inputs.
Let us denote intervals over matrices as $\boldsymbol{A} := [{A}^L, {A}^U] \subseteq \mathbb{R}^{n \times m}$ such that $A \in\boldsymbol{A} \implies {A}^L \leq A \leq {A}^U$.
We define the following interval matrix arithmetic operations\footnote{We note the absence of interval division, which poses challenges when the divisor interval contains zero; we will not require this operation in our subsequent analysis.}:
\begin{align*}
\text{Addition: } \  &A + B \in [\boldsymbol{A} \oplus\boldsymbol{B}] \ \forall A \in \boldsymbol{A}, B \in \boldsymbol{B}\\
\text{Subtraction: } \  &A - B \in [\boldsymbol{A} \ominus \boldsymbol{B}] \ \forall A \in \boldsymbol{A}, B \in \boldsymbol{B}\\
\text{Matrix multiplication: } \  &A\times B \in [\boldsymbol{A} \otimes \boldsymbol{B}] \ \forall A \in \boldsymbol{A}, B \in \boldsymbol{B}\\
\text{Elementwise multiplication: } \  &A\circ B \in [\boldsymbol{A} \odot \boldsymbol{B}] \ \  \forall A \in \boldsymbol{A}, B \in \boldsymbol{B}\\
\text{Scalar multiplication: } \  &\alpha A \in [\alpha \boldsymbol{A}] \ \forall A \in \boldsymbol{A}
\end{align*}
Each of these operations can be computed (or over-approximated) using standard interval arithmetic in at most 4x the computational cost of its non-interval counterpart.
For example, interval matrix multiplication can be computed efficiently using Rump or Nguyen's algorithms \citep{rump1999fast, nguyen2012efficient}.
We denote interval vectors as $\boldsymbol{a} := \left[{a}^L, {a}^U\right]$ with analogous operations.

\subsection{Interval Decomposition of Abstract Gradient Training.}

\begin{figure}
    \centering
    \includegraphics[width=\linewidth]{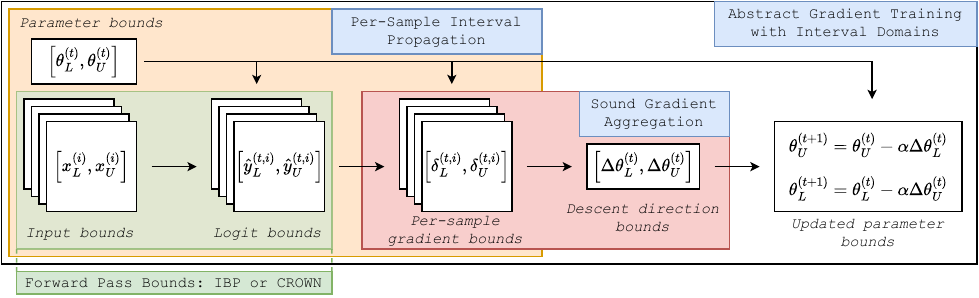}
    \caption{Summary of AGT with interval domains: (1) Forward pass bounds are computed using either interval bound propagation or interval-CROWN. (2) Per-sample gradient bounds are computed using interval backpropagation. (3) Descent direction bounds are computed using sound aggregation of per-sample gradients with respect to a given perturbation model. (4) The updated parameter interval is computed using interval arithmetic.}
    \label{fig:interval_gradient_training}
\end{figure}

Interval domains provide a natural representation for model parameters, perturbed data, model activations, and gradient updates within the Abstract Gradient Training framework. 
We begin the training process with the parameter interval $\boldsymbol{\theta^{(0)}}=[\theta^{(0)}, \theta^{(0)}]$, and seek to compute a valid parameter-space interval, $\boldsymbol{\theta^{(t)}}$, for each subsequent training iteration.

Each iteration updates model parameters based on the gradient of the loss function. In standard SGD, the parameter update at iteration $t$ given by \eqref{eq:sgd} can be re-written as:
\begin{alignat}{2}
\hat{y}^{(t, i)} &= f\left({x}^{(i)}, \theta^{(t-1)}\right), \quad \delta^{(t, i)} &&= \nabla_\theta\mathcal{L}\left(\hat{y}^{(t, i)}, {y}^{(i)}\right),\\
\Delta \theta^{(t)} &= \frac{1}{|\mathcal{I}^{(t)}|}\sum\limits_{i \in \mathcal{I}^{(t)}} \delta^{(t, i)}, \quad \theta^{(t)} &&= \theta^{(t-1)} - \alpha \Delta \theta^{(t)},
\end{alignat}
where again the index set $\mathcal{I}^{(t)} \subset [N]$ captures the data ordering (i.e. the indices of the data-points included in the $t$-th batch). We will refer to $\Delta \theta^{(t)}$ as the average gradient, or \textit{descent direction}, computed over the training batch $\mathcal{B}^{(t)}$.
With interval domains, our objective is to compute sound interval enclosures over each of the operations above:
\begin{alignat}{2}
\hat{y}^{(t, i)} &= \interval{\hat{y}^{(t, i)}}, \quad
\delta^{(t, i)} &&= \interval{\delta^{(t, i)}}, \\
\Delta \theta^{(t)} &= \interval{\Delta \theta^{(t)}}, \quad
\theta^{(t)} &&= \interval{\theta^{(t)}}.
\end{alignat}
We address this challenge via a two-stage approach, as illustrated in Figure~\ref{fig:interval_gradient_training}:
\begin{enumerate}
    \item \textbf{Per-sample gradient bounds:} We first derive valid interval bounds for each \textit{per-sample} gradient term $\delta^{(t, i)}$, by propagating bounds through both the forward and backward passes of the neural network. We instantiate two strategies for this step: interval bound propagation, and an extension of the CROWN algorithm adapted to the interval-weight setting.
    \item \textbf{Sound aggregation:} We then apply a \textit{sound aggregation mechanism} to bound the summation of these individual gradient terms, taking into account the specific perturbation model $\mathcal{T}$, and ultimately producing a sound interval enclosure of the descent direction $\Delta \theta^{(t)}$. This step ensures that the resulting bounds reflect the fact that \emph{up to} $n$ data points may be altered.
\end{enumerate}
The remainder of this section specifies the exact computations involved in each of these steps.

\subsection{Computing Per-Sample Gradient Bounds}\label{sec:per-sample-propagation}

\begin{figure}[t]
    \centering
    \includegraphics[width=0.9\linewidth]{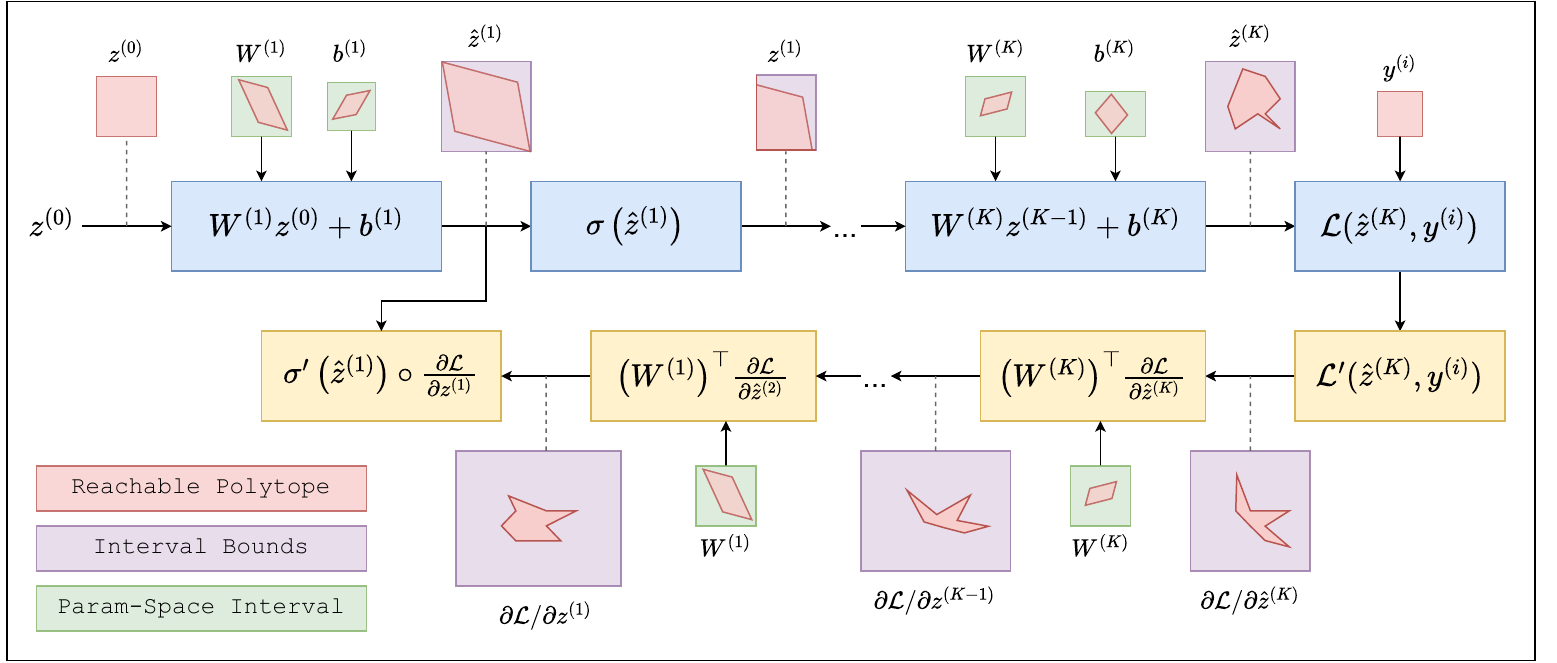}
    \caption{Bounding per-sample gradients using interval bound propagation. Top: Interval propagation through the forward pass, with respect to both input and parameter intervals. Bottom: Interval propagation through the backward pass of the network. Gradients with respect to model parameters are omitted for clarity.}
    \label{fig:ibp}
\end{figure}

This section describes interval propagation for computing sound interval bounds on per-sample gradients.
Let $\delta^{(t, i)}$ denote the gradient of the loss function $\mathcal{L}$ with respect to the model parameters $\theta$, evaluated for the $i$-th sample at the $t$-th training iteration:
\begin{equation}
    \delta^{(t, i)} = \nabla_\theta \mathcal{L}\left(f\left({x}^{(i)}, \theta^{(t-1)}\right), y^{(i)}\right).
\end{equation}

Our objective is to derive sound interval bounds for $\delta^{(t, i)}$ that capture all perturbations introduced by the perturbation model $\mathcal{T}$.
Depending on the specific choice of $\mathcal{T}$, these bounds must be sound with respect to:
\begin{itemize}
    \item Only the parameter interval $\boldsymbol{\theta^{(t-1)}}$ (e.g., when considering sample removal/substitution).
    \item Both the parameter interval $\boldsymbol{\theta^{(t-1)}}$ and intervals representing perturbations of the input $x^{(i)} \in \boldsymbol{x^{(i)}}$ and label $y^{(i)} \in \boldsymbol{y^{(i)}}$ (e.g., when considering bounded input and label modifications).
\end{itemize}
For generality, we focus on the second, more comprehensive case.
The first case can be treated as a special instance by setting $\boldsymbol{x^{(i)}}$ and $\boldsymbol{y^{(i)}}$ to zero-width intervals, effectively eliminating their contribution to the bounds and reducing the problem to one with parameter-space perturbations only.

\begin{remark}
    While the bounded perturbation model allows for general $\ell_p$- norm perturbations, in this section we focus exclusively on their over-approximations using the $\ell_\infty$ norm. This approach provides a sound, though potentially looser, bound. For instance, the interval over the input feature space is expressed as $\boldsymbol{x^{(i)}} = \left[x^{(i)} - \epsilon, x^{(i)} + \epsilon \right]$.
\end{remark}

To compute sound gradient bounds, we decompose the gradient computation process into three independent stages, which we detail below.

\paragraph{1) Bounding the Forward Pass.}
Computing bounds on the forward pass of a neural network is straightforwardly defined using interval arithmetic.
The parameter interval $\boldsymbol{\theta^{(t-1)}}$ defines intervals on the weights and biases of the network, represented as $W^{(k)} \in \boldsymbol{W^{(k)}}$ and $b^{(k)} \in \boldsymbol{b^{(k)}}$ for each layer $k = 1, \dots, K$.
Given these interval parameters, and an interval $\boldsymbol{z^{(0)}} = \boldsymbol{x}$ on the (perturbed) input to the network, we recursively propagate bounds through the network layers.
The pre-activation and post-activation interval bounds at each layer are computed as:
\begin{equation}
    \boldsymbol{\hat{z}^{(k)}} = \boldsymbol{W^{(k)}} \otimes \boldsymbol{z^{(k-1)}} \oplus \boldsymbol{b^{(k)}}, \quad \boldsymbol{z^{(k)}} = \sigma \left(\boldsymbol{\hat{z}^{(k)}}\right).
\end{equation}
Here, $\boldsymbol{\hat{z}^{(k)}}$ represents the interval bounds on the pre-activation values, while $\boldsymbol{z}^{(k)}$ contains the bounds on the post-activation values.  
For a monotonic activation function $\sigma$, interval propagation is performed by independently applying $\sigma$ to the lower and upper bounds of its input, i.e., $\sigma\left(\left[{a}^L, {a}^U\right]\right) = \left[\sigma\left({a}^L\right), \sigma\left({a}^U\right)\right]$.

\begin{remark}
We note that it is possible to employ more precise (but computationally heavier) approaches in this step to compute tighter intervals over the forward pass.
While this step differs from standard adversarial robustness certification due to the inclusion of intervals over the model parameters, many existing methods can be adapted to this more general setting.
As an example, we provide an extension of the well-known CROWN algorithm to handle interval-valued parameters in Appendix~\ref{app:crown}.
\end{remark}

\paragraph{2) Bounding the Loss Function.}
The first step of the backward pass involves computing the partial derivative of the loss $\mathcal{L}(\hat{y}, y)$ with respect to the network logits $\hat{y}$.
The interval bounds must be valid with respect to both the logit interval $\boldsymbol{\hat{y}} = \boldsymbol{\hat{z}^{(K)}}$ (obtained from the forward pass) and the label interval $\boldsymbol{y}$ (given by the perturbation model).

For common loss functions such as cross-entropy and squared error loss, these gradients can be explicitly derived, allowing interval arithmetic to compute sound enclosures over the required terms.
Consider the squared error loss, defined as $\mathcal{L}\left(\hat{y}, y\right) = \|\hat{y} - y \|^2_2$, which has a partial derivative given by ${\partial \mathcal{L}} / {\partial \hat{y}} = 2 (\hat{y} - y)$.
Applying interval arithmetic to this gradient expression yields a sound enclosure for the loss gradient:
\begin{equation}
   \boldsymbol{\frac{\partial \mathcal{L}}{\partial \hat{y}}} = 2 (\boldsymbol{\hat{y}} \ominus \boldsymbol{y}).
\end{equation}

Computing interval bounds on derivatives of other loss functions, e.g., cross-entropy loss, requires propagation through additional non-linear terms including the softmax function.
Interval propagation for the cross-entropy and hinge loss functions are detailed in Appendix~\ref{app:ibp_loss_fn}.

\paragraph{3) Bounding the Backward Pass.} 
Finally, to establish interval bounds on the full backward pass of the network, we extend the interval arithmetic-based approach of \citet{wicker2022robust}, which provides bounds on derivatives of the form ${\partial \mathcal{L}} / {\partial z^{(k)}}$.
Our extension additionally computes sound bounds on gradients with respect to the network parameters (i.e., ${\partial \mathcal{L}} / {\partial W^{(k)}}$ and ${\partial \mathcal{L}} / {\partial b^{(k)}}$).

For a feed-forward neural network, the backward pass using standard back-propagation follows the recursive update rule:
\begin{align}
\frac{\partial \mathcal{L}}{\partial z^{(k-1)}} &= \left(W^{(k)}\right)^\top \frac{\partial \mathcal{L}}{\partial \hat{z}^{(k)}}, \quad \frac{\partial \mathcal{L}}{\partial \hat{z}^{(k)}} = \sigma'\left(\hat{z}^{(k)}\right) \circ \frac{\partial \mathcal{L}}{\partial z^{(k)}} \\
\frac{\partial \mathcal{L}}{\partial W^{(k)}} &= \frac{\partial \mathcal{L}}{\partial \hat{z}^{(k)}} \left(z^{(k-1)}\right)^\top, \quad \frac{\partial \mathcal{L}}{\partial b^{(k)}} = \frac{\partial \mathcal{L}}{\partial \hat{z}^{(k)}},
\end{align}
where $\sigma'(\cdot)$ denotes the derivative of the activation function, and $\circ$ represents element-wise multiplication.
To bound the backward pass, we leverage the following intervals:
\begin{enumerate}
   \item Interval bounds on intermediate activations $z^{(k)}$ and pre-activations $\hat{z}^{(k-1)}$ obtained from the forward pass.
   \item Bounds on the initial loss gradient, $\frac{\partial \mathcal{L}}{\partial \hat{z}^{(K)}}$, computed using the interval propagation procedure described previously.
\end{enumerate}

Using these components, we derive sound interval bounds for the gradients through the following interval matrix operations:
\begin{align}
\boldsymbol{\frac{\partial \mathcal{L}}{\partial z^{(k-1)}}} &= \left(\boldsymbol{W^{(k)}}\right)^\top \otimes \boldsymbol{\frac{\partial \mathcal{L}}{\partial \hat{z}^{(k)}}}, \quad \boldsymbol{\frac{\partial \mathcal{L}}{\partial \hat{z}^{(k)}}} = \sigma'\left(\boldsymbol{\hat{z}^{(k)}}\right) \odot \boldsymbol{\frac{\partial \mathcal{L}}{\partial z^{(k)}}} \\
\boldsymbol{\frac{\partial \mathcal{L}}{\partial W^{(k)}}} &= \boldsymbol{\frac{\partial \mathcal{L}}{\partial \hat{z}^{(k)}}} \otimes \left(\boldsymbol{{z}^{(k-1)}}\right)^\top, \quad \boldsymbol{\frac{\partial \mathcal{L}}{\partial b^{(k)}}} = \boldsymbol{\frac{\partial \mathcal{L}}{\partial \hat{z}^{(k)}}},
\end{align}
where all interval operations follow previously defined rules.
Interval propagation through the activation function derivative $\sigma'(\cdot)$ depends on the specific activation function employed. For ReLU networks, the derivative is the Heaviside function, for which interval bounds can be soundly propagated as $\sigma'([a^L, a^U]) = [\sigma'(a^L), \sigma'(a^U)]$.

The resulting gradient intervals provide a complete interval enclosure of all gradients of the model for all $W^{(k)} \in \boldsymbol{W^{(k)}}$, $b^{(k)} \in \boldsymbol{b^{(k)}}$ given a single sample $\tilde{x} \in \boldsymbol{\tilde{x}}$, $\tilde{y} \in \boldsymbol{\tilde{y}}$.

\begin{figure}[t]
    \centering
    \includegraphics[trim={0 0 0.5cm 0}, width=0.9\linewidth]{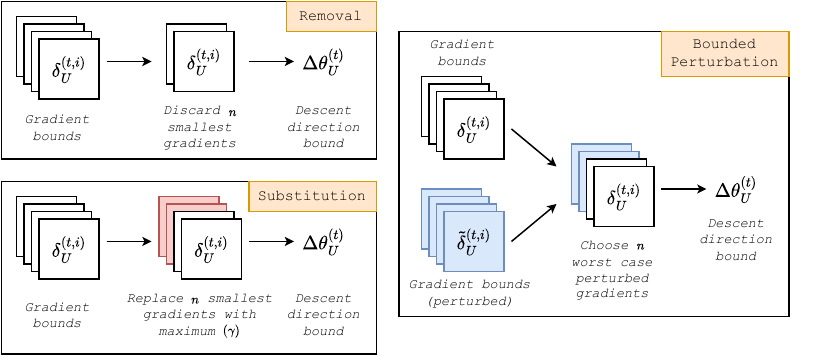}
    \caption{
    Sound aggregation of per-sample gradients for our three perturbation models.
    Each case illustrates the procedure for upper-bounding a single parameter index, with the full bounds obtained by independently repeating the process for each parameter index. The lower bound is not shown but follows analogous operations.
}\label{fig:aggregation}
\end{figure}

\subsection{Computing Descent Direction Bounds via Sound Aggregation}\label{sec:sound_agg}

Having established sound interval bounds on the per-sample gradients, we now address the challenge of computing tight interval bounds on the overall descent direction $\boldsymbol{\Delta \theta^{(t)}}$. This requires careful consideration of the aggregation process across the per-sample gradients, while accounting for the specific perturbation model $\mathcal{T}$.

The descent direction in standard SGD is computed as an average of individual gradients across the batch, $\Delta \theta^{(t)} = \sum_i\delta^{(t, i)} / {|\mathcal{B}^{(t)}|}$.
When considering an interval bound on this term, a naive approach would be to simply compute the interval sum of per-sample gradient bounds and divide by the batch size.
However, this approach yields overly conservative bounds, as it ignores other dependencies implied by the constraints of the perturbation model $\mathcal{T}$, such as being able to modify only up to $n$ training samples in the dataset.
Instead, we develop tailored procedures for bounding the descent direction that exploit the structure of specific perturbation models to obtain valid and tighter bounds.

Before presenting the aggregation procedures for each perturbation model, we establish the following notation:
\begin{itemize}
   \item Let ${\delta}^{(t, i)}_L$ and ${\delta}^{(t, i)}_U$ be bounds capturing the effect of perturbations in previous iterations, satisfying:
   ${\delta}^{(t, i)}_L \leq \delta^{(t, i)} \leq {\delta}^{(t, i)}_U$, $\forall \delta^{(t, i)} \in \left\{ \nabla_{{\theta}} \mathcal{L} \left( f\big(x^{(i)}, \tilde{\theta}^{(t-1)}\big), y^{(i)} \right) \mid \tilde{\theta}^{(t-1)} \in \boldsymbol{\theta^{(t-1)}} \right\}$.
   \item Let $\tilde{\delta}^{(t, i)}_L$ and $\tilde{\delta}^{(t, i)}_U$ be bounds on the \textit{combined} effect of previous perturbations and worst-case bounded perturbations of the current batch, satisfying:
   $\tilde{\delta}^{(t, i)}_L \leq \tilde{\delta}^{(t, i)} \leq \tilde{\delta}^{(t, i)}_U$ for all $\tilde{\delta}^{(t, i)}$ in
 \begin{equation*}
 \left\{ \nabla_{{\theta}} \mathcal{L} \left( f\big(\tilde{x}, \tilde{\theta}^{(t-1)}\big), \tilde{y} \right) \mid
      \tilde{\theta}^{(t-1)} \in \boldsymbol{\theta^{(t-1)}},
      \|x^{(i)} - \tilde{x} \|_p \leq \epsilon,
      \|y^{(i)} - \tilde{y} \|_q \leq \nu
 \right\}.
 \end{equation*}
\end{itemize}
These bounds can be computed using the interval propagation procedure outlined in the previous section. For the second case, we over-approximate the norm constraints using the intervals $\boldsymbol{\tilde{x}} =  [x^{(i)} - \epsilon, x^{(i)} + \epsilon]$ and $\boldsymbol{\tilde{y}} = [y^{(i)} - \nu, y^{(i)} + \nu]$.

Throughout our analysis, we assume that the perturbation model $\mathcal{T}$ can act independently on each batch $\mathcal{B}^{(t)}$.
While this is a relaxation compared to $\mathcal{T}$ acting once on the entire dataset, it remains sound and facilitates tractable computation of bounds.
Consequently, our guarantees hold for up to $n$ modified points per batch. However, any certificates defined over the entire dataset remain valid only for up to $n$ modified points per dataset, reflecting the worst-case scenario in which all modifications occur within a single batch.

\begin{remark}
In poisoning scenarios, if we adopt a more relaxed attack model -- assuming the adversary lacks access to data ordering (i.e., is not an online adversary as in \cite{zhang2020online}) and that poisoned data is uniformly distributed across batches -- we can specify our guarantees with respect to a ``safe'' proportion of poisoned samples per batch by applying statistical bounds on the likelihood of exceeding a certain number of poisoned samples within any batch. This approach yields less conservative, though probabilistic, guarantees compared to the worst-case deterministic bounds presented here.
\end{remark}

A summary of our sound aggregation mechanisms for each perturbation model is shown in Figure~\ref{fig:aggregation}. Below we detail the specifics of each mechanism.

\subsubsection{Bounding the Descent Direction under Removal.}
The most straightforward perturbation model is the \textit{removal} of $n$ training samples from the batch.
To bound this case, we observe that for any particular parameter index, the average gradient is maximized by removing the $n$ samples with the smallest gradients, and minimized by removing the $n$ samples with the largest gradients.

The effect of the removal perturbation model can be bounded by applying this reasoning at \textit{each parameter index} independently.
This represents an over-approximation of the actual effects, since any realizable training data perturbation must select the same $n$ points to remove across all parameter indices.
However, this trade-off between parameter independence and computational efficiency is necessary when working within the interval domain.

The following theorem formalizes how to compute bounds on the descent direction under the data removal perturbation model:
\begin{theorem}\label{thm:aggregation_unlearning}
   Given a nominal batch $\mathcal{B}^{(t)} = \left\{\left(\tilde{x}^{(i)}, \tilde{y}^{(i)} \right)\right\}_{i=1}^b$ of size $b$, an interval parameter domain $\boldsymbol{\theta^{(t-1)}}$, the descent direction 
   \begin{equation}
       \Delta \theta^{(t)} = \frac{1}{|\mathcal{\tilde{B}}^{(t)}|}\sum\limits_{\left(\tilde{x}^{(i)}, \tilde{y}^{(i)}\right) \in \mathcal{\tilde{B}}^{(t)}}        
       \nabla_\theta  \mathcal{L} \left(f\big(\tilde{x}^{(i)}, \theta^{(t-1)} \big), \tilde{y}^{(i)}\right)
   \end{equation}
   is bounded element-wise by
   \begin{equation}
       \Delta \theta_L^{(t)} = \frac{1}{b - n}\left(\underset{b - n}{\operatorname{SEMin}}
       \left\{
       {\delta^{(t, i)}_L}\right\}_{i=1}^b
       \right), \quad
       \Delta \theta_U^{(t)} = \frac{1}{b - n} \left(\underset{b - n}{\operatorname{SEMax}}
       \left\{
       {\delta^{(t, i)}_U}\right\}_{i=1}^b
       \right)
   \end{equation}
   for any perturbed batch $\mathcal{\tilde{B}}^{(t)}$ derived from $\mathcal{B}^{(t)}$ by removing up to $n$ data points.
\end{theorem}
In this theorem, the operations $\operatorname{SEMax}_a$ and $\operatorname{SEMin}_a$ correspond to taking the sum of the top-$a$ and bottom-$a$ elements at each index, sorted by magnitude, respectively.

\subsubsection{Bounding the Descent Direction under Substitution.}
Extending this approach to the case of data substitution introduces a challenge in bounding the descent direction.
Specifically, the inclusion of an arbitrary data point can have an unbounded effect on the model's gradient, making it impossible to enclose the descent direction within intervals without modifying the SGD update step.
To address this, we introduce a \textit{clipped} SGD update, defined as
\begin{equation}
    \Delta \theta^{(t)} = \frac{1}{|\mathcal{B}^{(t)}|}\sum\limits_{ (\tilde{x}^{(i)}, \tilde{y}^{(i)}) \in \mathcal{B}^{(t)}} \operatorname{Clip}_\kappa \left[\nabla_\theta  \mathcal{L} \left(f\big(\tilde{x}^{(i)}, \theta^{(t-1)} \big), \tilde{y}^{(i)}\right)\right],
\end{equation}
where $\operatorname{Clip}_\kappa$ is a truncation operator that clamps all elements of its input to be between $-\kappa$ and $\kappa$, while leaving those within the range unchanged.
This allows consideration of \textit{arbitrary substitutions}, but the resulting reachable parameter sets are only valid with respect to the training process that applies the same clipping procedure to the nominal parameters.

\begin{remark}
Per-sample gradient clipping is widely used in machine learning to restrict the impact of individual training examples, such as in differentially private training \citep{abadi2016deep}. Unlike the commonly used $\ell_2$-norm clipping, here we opt for the truncation operator, as it is better suited for interval bound propagation.
\end{remark}

Since the clipping procedure is applied element-wise, we can conclude that the worst-case addition of $n$ datapoints can only have an effect of up to $\pm n \kappa$ on the sum of the gradients.
Specifically, we have the following result:
\begin{theorem}\label{thm:aggregation_privacy}
    Given a nominal batch $\mathcal{B}^{(t)} = \left\{ \left( x^{(i)}, y^{(i)} \right) \right\}_{i=1}^b$, and an interval parameter domain $\boldsymbol{\theta^{(t-1)}}$, the descent direction
    \begin{equation}
        \Delta \theta^{(t)} = \frac{1}{|\mathcal{\tilde{B}}^{(t)}|}\sum\limits_{\left({x}^{(i)}, {y}^{(i)}\right) \in \mathcal{\tilde{B}}^{(t)}}        
        \operatorname{Clip}_\kappa \left[\nabla_\theta  \mathcal{L} \left(f\big({x}^{(i)}, \theta^{(t-1)} \big), {y}^{(i)}\right)\right]
    \end{equation}
    is bounded element-wise by
    \begin{equation}
        \Delta \theta_L^{(t)} = \frac{1}{b}\left(\underset{b - n}{\operatorname{SEMin}}
        \left\{
        {\delta^{(t, i)}_L}\right\}_{i=1}^b - n \kappa \mathbf{1}
        \right), \quad
        \Delta \theta_U^{(t)} = \frac{1}{b} \left(\underset{b - n}{\operatorname{SEMax}}
        \left\{
        {\delta^{(t, i)}_U}\right\}_{i=1}^b + n \kappa \mathbf{1}
        \right)
    \end{equation}
    for any perturbed batch $\mathcal{\tilde{B}}^{(t)}$ derived from $\mathcal{B}^{(t)}$ removing up to $n$ data-points and adding up to $n$ arbitrary data-points.
\end{theorem}

\subsubsection{Bounding the Descent Direction under Bounded Perturbation.}

Finally, we analyze the effect of bounded perturbations to existing training data samples.  
We begin by noting that perturbing the $i$-th data point impacts our bounds by an additional contribution of $\left(\tilde{\delta}^{(t, i)}_U - {\delta}^{(t, i)}_U\right)$ to the upper bound and $\left(\tilde{\delta}^{(t, i)}_L - {\delta}^{(t, i)}_L\right)$ to the lower bound.
By independently selecting the worst-case $n$ contributions at each parameter index, we derive an efficient method for computing a valid interval over the overall descent direction.
Formally, we state the following result:  

\begin{theorem}\label{thm:aggregation_poisoning}
    Given a nominal batch $\mathcal{B}^{(t)} = \left\{ \left( x^{(i)}, y^{(i)} \right) \right\}_{i=1}^b$, and an interval parameter domain $\boldsymbol{\theta^{(t-1)}}$, the descent direction
    \begin{equation}
        \Delta \theta^{(t)} = \frac{1}{|\mathcal{\tilde{B}}^{(t)}|}\sum\limits_{\left(\tilde{x}^{(i)}, \tilde{y}^{(i)}\right) \in \mathcal{\tilde{B}}^{(t)}}
        \nabla_\theta  \mathcal{L} \left(f\big({x}^{(i)}, \theta^{(t-1)} \big), {y}^{(i)}\right)
    \end{equation}
    is bounded element-wise by
    \begin{align}
    \Delta \theta^L &= \frac{1}{b} \left( \underset{n}{\operatorname{SEMin}} \left\{ \tilde{\delta}^{(t, i)}_L - {\delta}^{(t, i)}_L \right\}_{i=1}^b + \sum\limits_{i=1}^b {\delta}^{(t, i)}_L \right),\\
    \Delta \theta^U &= \frac{1}{b} \left( \underset{n}{\operatorname{SEMax}} \left\{ \tilde{\delta}^{(t, i)}_U - {\delta}^{(t, i)}_U \right\}_{i=1}^b + \sum\limits_{i=1}^b {\delta}^{(t, i)}_U \right).
    \end{align}
    for any batch $\mathcal{\tilde{B}}^{(t)}$ derived from $\mathcal{{B}}^{(t)}$ by bounded perturbation within $\mathcal{T}^{n, p, \epsilon, q, \nu}_{\text{bounded}}$ given by \eqref{eq:bound_adv}.
\end{theorem}
As before, the above result soundly bounds the descent direction for all parameters, with the update being applied independently at each parameter index.
This approach, while computationally efficient for propagating interval enclosures, introduces a potentially loose over-approximation, as the $n$ worst-case points for a given parameter index are unlikely to be the same as those for the parameters at all other indices.

\subsection{Summary of Abstract Gradient Training with Interval Domains}

\subfile{algorithm}

The instantiation of Abstract Gradient Training using interval arithmetic facilitates computationally efficient certification of large-scale training pipelines.
However, this efficiency comes at the cost of potentially significant over-approximation of the reachable parameter space at each training iteration, which may reduce the tightness of the resulting guarantees.
Algorithm~\ref{alg:abstractgradtrain} summarizes the key steps of this approach in pseudo-code; we now outline additional computational considerations that arise in practice.

\textbf{Computational Complexity.}
The computational cost of AGT with interval domains is dependent on the forward-pass bounding method.
Simple IBP incurs a cost roughly four times that of standard forward and backward passes.
Similarly, our CROWN-based bounds increase the cost by a factor of four compared to the original CROWN algorithm, which has a complexity of $\mathcal{O}(m^2 n^3)$ for an $m$-layer network with $n$ neurons per layer and $n$ outputs \citep{zhang2018efficient}.
The $\operatorname{SEMin}/\operatorname{SEMax}$ operations, while $\mathcal{O}(b)$ per index, are efficiently parallelizable on GPUs.
Empirically, we observe that AGT with interval domains typically adds less than a 2x runtime overhead to standard training.

\textbf{Limitations.}
While AGT with interval domains provides valid parameter space bounds for any gradient-based training algorithm, the tightness of these bounds varies based on architecture, hyperparameters, and training procedures.
Notably, the algorithm assumes simultaneous worst-case perturbation at each parameter index, in addition to our relaxation of worst-case batch perturbations.
This approach, while computationally efficient for propagating interval enclosures, introduces a potentially loose over-approximation, as the $n$ worst-case perturbed points for any parameter index are unlikely to be the same for all indices.
Consequently, we found that achieving meaningful guarantees with this approach often necessitates larger batch sizes or fewer training epochs. Furthermore, loss functions like multi-class cross-entropy exhibit loose interval relaxations, leading to weaker guarantees compared to those for regression or binary classification.

%% file: sections/algorithm.tex
\begin{algorithm*}[tb]
   \caption{\textsc{Abstract Gradient Training with Interval Domains}}\label{alg:abstractgradtrain}
   \small{
\begin{algorithmic}[1]
\STATE {\bfseries input:} $f$ - model, $\mathcal{D}$ - dataset, $\{\mathcal{I}\}_{t=1}^T$ - data ordering, $\theta^{(0)}$ - parameter initialization, $\alpha$ - learning rate, $\mathcal{T}$ - perturbation model.
\STATE {\bfseries output:} {$\intervalshort{\theta^{(T)}}$ - valid parameter space domain}
   \STATE $\left[\theta^{(0)}_L, \theta^{(0)}_U\right] \gets \left[\theta^{(0)}, \theta^{(0)}\right]$  
   \FOR {$t = 1, \dots, T$}
   \FOR {$i \in \mathcal{I}^{(t)}$}
    \STATE{$\intervalshort{\hat{y}^{(t, i)}} \gets$ \textsc{bound\_forward\_pass}($x^{(i)}, \intervalshort{\theta^{(t)}}, \mathcal{T}$) \hfill $\triangleright$ IBP \S\ref{sec:per-sample-propagation} or CROWN \S\ref{app:crown}}
    \STATE{$\intervalshort{{\delta}^{(t, i)}} \gets$ \textsc{bound\_backward\_pass}($\intervalshort{\hat{y}^{(t, i)}}, \intervalshort{\theta^{(t)}}, y^{(i)}, \mathcal{T}$) \hfill $\triangleright$ IBP \S\ref{sec:per-sample-propagation}}
   \ENDFOR
    \STATE {$\intervalshort{\Delta\theta^{(t)}} \gets$ \textsc{sound\_aggregation}$(\{\intervalshort{{\delta}^{(t, i)}}\}_i, \mathcal{T})$ \hfill $\triangleright$ Aggregation Mechanism \S\ref{sec:sound_agg}}
    \STATE {$\intervalshort{\theta^{(t)}} \gets [\theta^{(t-1)}_L - \alpha \Delta \theta^{(t)}_U ,\theta^{(t-1)}_U - \alpha \Delta \theta^{(t)}_L]$ \hfill $\triangleright$ Parameter Interval Update}
   \ENDFOR
   \STATE \textbf{return} $\intervalshort{\theta^{(T)}}$
\end{algorithmic}
}
\end{algorithm*}

%% file: sections/07_optimization.tex
\section{Tight Analysis via Optimization-Based Certification}\label{sec:optimization}
In the previous section, we established a sound and computationally efficient approach for bounding the training dynamics using interval propagation.
While efficient, this approach introduces over-approximations at each iteration, which can lead to loose or even vacuous certification guarantees.
To achieve a tighter certification, we revisit the optimization-based formulation of AGT, aiming for a more precise characterization of training perturbations.

This section begins with a mixed-integer programming (MIP) formulation that provides an exact representation of our abstract training dynamics.
We then explore convex and linear relaxations that balance computational efficiency and certification tightness.
Finally, we introduce decomposition strategies that partition the AGT problem into smaller sub-problems, potentially enabling more scalable certification for larger-scale training settings.

\subsection{Mixed-Integer Programming}

To obtain an exact representation of our abstract training dynamics, we formulate the AGT constrained optimization problem as a \textbf{mixed-integer program (MIP)}.
Unlike interval propagation, which provides a single lower and upper bound for each parameter,  MIP enables an exact characterization of the worst-case perturbations by explicitly encoding discrete constraints and non-linear dependencies over the entire training trajectory.
However, this precision comes at the cost of increased computational complexity, with mixed-integer programs being NP-Hard, in general.

Mixed-integer programming is a powerful optimization framework that extends linear programming by incorporating both continuous and discrete variables.
A standard mixed-integer program takes the form:
\begin{alignat}{2}
\min \ & h(a, b) \\
\text{s.t. } & g_j(a, b) \leq 0, \qquad  j = 1, \dots, m, \\
& a \in \mathbb{R}^{n_c}, b \in \mathbb{Z}^{n_i}
\end{alignat}
where $a$ represents $n_c$ continuous decision variables, $b$ represents $n_i$ integer decision variables, and $g_i(\cdot)$ are $m$ constraint functions that may include both linear and non-linear components.
The inclusion of integer variables allows MIP to model complex non-convex decision problems, such as those including neural networks, logical constraints or piecewise functions.

The structure of the objective and constraint functions defines the specific class of optimization problem.
When the objective function $h$ is linear and the constraints $g_i$ include quadratic or bilinear terms in the decision variables, the problem falls under the category of a \textit{mixed-integer quadratically constrained program (MIQCP)}.
If both the objective function and all constraints are purely linear, the formulation simplifies to a \textit{mixed-integer linear program (MILP)}, which generally allows for more efficient solving methods.
Finally, if the integer variables are relaxed to continuous domains, then the problem may be a \textit{quadratically constrained program (QCP)} or \textit{linear program (LP)}.

In the context of AGT, MIQCPs offer the greatest expressive power, allowing for exact characterization of the training dynamics.
However, MIQCPs often require the most expensive solvers, whereas relaxations can provide a practical trade-off between precision and efficiency.

\subsection{Mixed-Integer Quadratically Constrained Representations of AGT}

To leverage MIP for AGT certification, we must express the constraints governing training perturbations, model structure, and parameter updates in a form compatible with mixed-integer programming.
This requires reformulating key components of the training process, such as activation functions, gradient computations, and adversarial modifications, using mixed-integer quadratic constraints.

\paragraph{Relation to previous works.} The formulations in this section are closely related to those of \cite{lorenz2024bicert}, which introduced a MIQCP formulation for certification against data poisoning.
While that work also considers training data perturbations, the forward pass, and gradient computations as mixed-integer programs, our approach offers a more comprehensive framework.
Specifically, while \cite{lorenz2024bicert} consider only bounded perturbations to all training examples, we expand the scope to a general set of perturbation models.
Moreover, their formulation is restricted to single training iteration per optimization problem, whereas our methodology considers the entire training procedure as a unified optimization problem.

\subsubsection{Dataset and Perturbation Models}
In our framework, training data perturbations arise from adversarial modifications or constraints imposed by differential privacy and machine unlearning.
To certify properties of the model with respect to these perturbation models, we must encode these modifications as feasible constraint sets.
We provide an explicit mixed-integer quadratic formulation for each of the perturbation models below:

\paragraph{Bounded Feature and Label Perturbation.} An adversary is allowed to modify a subset of up to $n$ training samples within predefined norms (here we present the formulation for the $p=\infty$ and $q=0$ norms).
To begin, we add a binary indicator variable $s \in \{0, 1\}^{|\mathcal{D}|}$ to represent the adversary's choice of poisoning targets.
Then, the feature and label space data poisoning constraints can be represented by the following linear constraints:
\begin{equation}
    \begin{array}{cl}
        x^{(i)} - \epsilon s_i \leq \tilde{x}^{(i)} \leq x^{(i)} + \epsilon s_i, & i = 1, \dots, |\mathcal{D}|\\
        y^{(i)} (1 - s_i) \leq \tilde{y}^{(i)} \leq y^{(i)} (1 - s_i) + s_i, & i = 1, \dots, |\mathcal{D}|\\
        \tilde{x}^{(i)} \in \mathbb{R}^{n_x}, \tilde{y}^{(i)} \in \{0, 1\} & i = 1, \dots, |\mathcal{D}|\\
        \sum\limits_{i=1}^{|\mathcal{D}|} s_i \leq n, & \\
    \end{array}
\end{equation}
The set of constraints above results in $s_i = 0 \implies (\tilde{x}^{(i)}, \tilde{y}^{(i)}) = ({x}^{(i)}, {y}^{(i)})$, while $s_i = 1$ indicates that the $i$-th training sample may be poisoned.
The above formulation covers binary classification only, but may be adapted to suit regression and multi-class problems.

\paragraph{Removal of Training Data.} We can encode for removal of training data similarly to the above, by again introducing an indicator variable $s \in \{0, 1\}^{|\mathcal{D}|}$ indicating the exclusion of a particular training sample.
Then, by adding the following constraint to each parameter update step, we can encode for the removal of $n$ or fewer training points:
\begin{equation}
    \sum\limits_{i=1}^{|\mathcal{D}|} s_i \leq n, \qquad \theta^{(t)} = \theta^{(t-1)} - \frac{\alpha}{|\mathcal{I}^{(t)}| - \sum_{i \in \mathcal{I}^{(i)}} s_i}\sum\limits_{ i \in \mathcal{I}^{(i)}} (1 - s_i) \delta^{(t, i)}, \qquad  t\in [T].
\end{equation}
Here $s_i = 1$ denotes the exclusion of the $i$-th training point from all training batches.

\paragraph{Substitution of Training Data.}
Finally, the substitution of existing training data by arbitrary training data can be handled by considering some bounded domain $\mathcal{X}^a$ from which the substituted data may be drawn.
This assumes the existence of such a set, though in practice such a set can be constructed by imposing bounds on the allowable feature space.
We can encode the added data through additional variables $\tilde{x}^{(j)} \in \mathcal{X}^a, \tilde{y}^{(j)} \in \{0, 1\}$ for $j \in 1, \dots, n$. The appropriate forward and backward passes encoded via the model constraints discussed below to obtain variables for the gradients of the model, denoted $\tilde{\delta}^{(t, j)}$.
Then, similar to the data removal case above we introduce binary decision variables $s \in \{0, 1\}^{|\mathcal{D}|}$ indicating the inclusion of each original data-point in the dataset, and $\tilde{s} \in \{0, 1\}^{n}$ indicating the index in the dataset which the perturbed point replaces.
The substituted data can then be expressed with the following constraints on the parameter update:
\begin{equation}
    \sum\limits_{i=1}^{|\mathcal{D}|} s_i =  \sum\limits_{j=1}^{n} \tilde{s}_j \leq n, \quad \theta^{(t)} = \theta^{(t-1)} - \frac{\alpha}{|\mathcal{I}^{(t)}|}\left[\sum\limits_{ i \in \mathcal{I}^{(i)}} (1 - s_i) \delta^{(t, i)} + \sum\limits_{j=1}^n \tilde{s}_j\tilde{\delta}^{(t, j)}\right], \quad  t\in [T].
\end{equation}
This formulation requires that when $s_i = 0$, the original training sample remains unchanged, while when $s_i = 1$, the sample is substituted with a new valid point from the given domain, with $n$ controlling the maximum number of substitutions allowed.

\subsubsection{Encoding the Forward Pass}

Representation of neural network models as mixed-integer programs is well established within the certification literature \citep{huchette2023deep}.
By encoding activation functions and linear layers as mixed-integer constraints, MIP formulations enable precise characterization of a network’s behavior under adversarial perturbations.
Unlike standard certification settings, in our case the network parameters $\theta=\left\{(W^{(i)}, b^{(i)})\right\}_{i=1}^K$ are also decision variables, leading to quadratic constraints.
In this work, we focus on the ReLU activation, which is representable as a mixed-integer linear program.
Several such MILP representations exist, such as those based on variable partitions \citep{tsay2021partition} or the convex hull \citep{anderson2020strong}, but here we present the simplest formulation, known as the `Big-M' formulation.
The $k$-th layer in the network, given by $z^{(k)} = \operatorname{ReLU}\left(W^{(k)} z^{(k-1)} + b^{(k)}\right)$, can be exactly represented as
\begin{align}
z^{(k)} &\geq W^{(k)} z^{(k-1)} + b^{(k)}, \label{eq:bigm1}\\
z^{(k)} &\leq W^{(k)} z^{(k-1)} + b^{(k)} - M s, \label{eq:bigm2}\\
0 \leq & \ z^{(k)} \leq Ms,\\
s \in &\{0, 1\}^{n_k}.
\end{align}
where $M$ is a sufficiently large constant and $s$ is a binary decision variable denoting the activation state of each neuron.
We highlight that constraints \eqref{eq:bigm1} and \eqref{eq:bigm2} are \textit{quadratic} constraints, since $W^{(k)}$ and $z^{(k-1)}$ are both decision variables.

\subsubsection{Encoding the Backward Pass}

To integrate the loss function into our optimization formulation, we must represent the partial derivative of the loss with respect to the model outputs.
Since not all loss functions are exactly MIQCP-representable, we focus only on the squared error and hinge loss here.
The more common cross entropy loss function cannot be encoded exactly, though relaxations are possible; we leave this possibility to future works.

\textit{Squared error loss:} The simplest loss function is again the squared error loss, which can be encoded via the linear constraint $\frac{\partial \mathcal{L}}{\partial \hat{y}} = 2 (\hat{y} - y)$.

\textit{Hinge loss:} The hinge loss, given by $\mathcal{L}(\hat{y}, y) = \max \{0, 1 - \hat{y}y\}$, can be re-written in terms of a ReLU activation function $\mathcal{L}(\hat{y}, y) = \operatorname{ReLU}(1 - \hat{y}y)$. Let the intermediate term by denoted by a variable $u=1 - \hat{y}y$. Then, we first encode $\operatorname{ReLU}(u)$ using the standard formulation above, noting that the gradient of the ReLU term is given by the binary variable $s$ present in the big-M formulation.
The gradient of the loss with respect to the logits $\hat{y}$ is then represented by the bilinear equality constraint $\partial \mathcal{L} / \partial \hat{y} = -ys$.

The backwards pass through the neural network is straightforward to encode using bilinear constraints.
We again note that the gradient of each ReLU activation is already encoded via the binary variable $s$ in the Big-M formulation above.

\subsubsection{Parameter Updates}
Finally, we can complete our formulation of the AGT training dynamics by including the updated model parameters.
For this constraint, we include a copy of the network parameters at each training iteration as a decision variable in our optimization problem.
With the exception of the removal perturbation model (which is discussed separately above), the linear constraints linking each iteration are simply
\begin{align}
  \theta^{(t)} &= \theta^{(t-1)} - \frac{\alpha}{|\mathcal{I}^{(t)}|}\sum\limits_{i \in \mathcal{I}^{(t)}} \delta^{(t, i)},\qquad t =1, \dots, T.
\end{align}
The decision variables $\delta^{(t, i)}$ are defined via the forward and backward pass constraints above.

\subsubsection{Choosing the MIP Objective Function}
The formulations outlined above allow us to encode the full training trajectory as a single MIQCP, capturing the constraints governing the evolution of model parameters.
To derive a final valid parameter-space domain for inference-time certification, we must compute bounds on the final parameter state, $\theta^{(T)}$.

There are multiple ways to define this final parameter domain.
A straightforward approach is to optimize for the minimum and maximum values of each parameter $\theta^{(T)}_j$ individually.
By setting the objective function to $\min / \max \theta^{(T)}_j$ for each index $j$ and solving the corresponding optimization problems, we obtain a valid interval enclosure for $\theta^{(T)}$.
This approach yields an interval domain that is significantly tighter than those obtained via interval bound propagation, as it fully accounts for dependencies among parameters during training.

For an even more refined representation of the final parameter space, we can compute the faces of a bounded polytope enclosing $\theta^{(T)}$.
Specifically, by selecting a set of face directions $A$ and maximizing each linear projection $A_i \theta^{(T)}$ as an objective function, we can construct a polyhedral domain of the form:
\begin{equation}
    \left\{\theta^{(T)} \mid A \theta^{(T)} \leq c\right\}
\end{equation}
This polytope representation provides a structured enclosure of the final parameter space, which can lead to tighter and more informative inference-time certification guarantees.

\subsection{Convex Relaxations of AGT}

While the MIQCP formulation provides a precise characterization of the training dynamics, its computational complexity can be prohibitive, especially for large-scale training pipelines.
The presence of integer variables in the optimization formulation combined with bilinear constraints significantly increases the difficulty of obtaining solutions.

To improve scalability while preserving the soundness of our certification guarantees, we explore relaxations that transform the original MIQCP into potentially more tractable linear optimization problems, which admit polynomial-time solutions.
By replacing non-convex constraints with convex outer approximations and relaxing integer constraints into continuous domains, we can formulate optimization problems that are more efficiently solvable using standard convex programming techniques.

\paragraph{Linearization of Integer Constraints.}
In our MIQCP formulation, integer variables play a crucial role in encoding discrete decisions, such as the selection of perturbed training samples or the activation behavior of piecewise-linear functions.
However, the inclusion of integer constraints significantly increases the computational complexity of solving the optimization problem.
Therefore, we investigate the \textit{continuous relaxation} of our formulation, which replaces any binary variables $s_i \in \{0, 1\}$ with a continuous counterpart $s_i \in [0, 1]$.
This relaxation weakens the formulation, for example by allowing the bounded threat model to partially poison more than $n$ points, but with the same total poisoning magnitude.

\paragraph{Linearization of Quadratic Constraints.}
Many of the constraints in our MIQCP formulation involve bilinear terms, such the product of model activations with network weight matrices. 
While this exactly encodes the model constraints, these bilinear terms introduce non-convexity into the optimization problem.
To improve tractability, we apply linearization techniques that replace bilinear terms by their over-approximate linear envelopes. 

A standard approach for linearizing a bilinear term $c = ab$, where $a \in [a_L, a_U]$ and $b \in [b_L, b_U]$ are bounded within known intervals, is to impose McCormick envelope constraints \citep{mccormick1976computability}:
\begin{align}
    c &\geq a_L b + a b_L - a_L b_L, \\
    c &\geq a_U b + a b_U - a_U b_U, \\
    c &\leq a_L b + a b_U - a_L b_U, \\
    c &\leq a_U b + a b_L - a_U b_L.
\end{align}
By applying this linearization to all bilinear terms in a formulation, we can form a convex outer approximation of the optimization problem, losing exactness but potentially leading to faster solve times.

\paragraph{Bounds Tightening.} 
The effectiveness of convex relaxations hinges on the tightness of the intermediate variable bounds used to construct the original formulation.
When linearizing mixed-integer or quadratic constraints, the gap between the mixed-integer optimal solution and its convex relaxation directly characterizes the ``tightness'' of the relaxation.
Big-M formulations are particularly sensitive to the choice of $M$, with excessively large values leading to weak relaxations \citep{tsay2021partition}.
Similarly, McCormick envelopes require tight intermediate bounds ($a_L$, $a_U$, $b_L$, $b_U$) to generate meaningful convex approximations \citep{mccormick1976computability}.
Recent advances have demonstrated substantial improvements in certification performance, particularly in deep problems, through iterative refinement of these bounds~\citep{sosnin2024scaling}.
In this work, we employ interval arithmetic for bounding intermediate variables, but more sophisticated bound tightening procedures, such as optimization-based bound tightening, represent promising directions for future work.

\subsection{Decomposition of the AGT Optimization Problem}

The formulations discussed above capture the entire training trajectory within a single large-scale optimization problem.
While this provides a precise characterization of the parameter evolution, the resulting problem requires $T$ copies of the model parameters, as well as a copy of the entire training dataset.
To improve scalability, we explore decomposition strategies that break the problem into smaller sub-problems that require only a subset of variables and constraints to be included.

\paragraph{Rolling-Horizon Decomposition.}
Instead of optimizing the entire training trajectory at once, rolling-horizon decomposition reformulates the AGT optimization problem as a sequence of smaller subproblems, each considering a limited time horizon. This approach mitigates the computational complexity and memory constraints associated with tracking the full trajectory while still capturing the dependencies between successive updates.

At a given training iteration $t$, we compute bounds on the parameters $\theta^{(t)}$ using only a fixed-length window of the most recent updates, rather than the entire training history. Specifically, we consider only the past $w$ iterations, where $w$ is a chosen window size.
In the next step, we compute bounds at iteration $t + p$, where $p \leq w$ is the window step size. This sliding window approach captures recent updates while ignoring older ones, whose influence on the current parameters naturally diminishes over time.

This decomposition strikes a balance between computational efficiency and bound tightness. By limiting the horizon of analysis, we enable incremental and scalable computation of certified bounds, making reachability analysis tractable over the course of training.

\paragraph{Sample-Wise Decomposition.}
While the rolling-horizon decomposition above can significantly reduce the size of the AGT sub-problems, it still requires consideration of an entire batch of training data within a single optimization problem.
For large models, this may still be computationally prohibitive, so another natural decomposition is a sample-wise decomposition, where the bounding problem is broken down across individual training samples, and per-sample gradients are bounded independently.
This requires the solution to many more sub-problems (at least an upper and lower bound for each parameter gradient at each training sample) but these smaller problems are solved independently and are readily parallelizable.
We leave this possibility to future works.

\subsection{Summary of Optimization-Based AGT}

\begin{table}[tbp]
\centering
\caption{Comparison of AGT formulation size for a neural network with $L$ layers and $W$ neurons per layer, trained on a dataset of size $N$ with $M$ features per entry for $T$ iterations of batch-size $B$.}
\label{tab:formulations}
\begin{tabular}{>{\raggedright\arraybackslash}p{1.5cm} >{\centering\arraybackslash}p{3.5cm} >{\centering\arraybackslash}p{4cm} >{\centering\arraybackslash}p{4cm}}
\toprule
\textbf{} & \textbf{\# Binary Variables} & \textbf{\# Continuous Variables} & \textbf{\# Quadratic Constraints} \\
\midrule
MIQCP & \small{$\mathcal{O}(WLBT + N)$} & \small{$\mathcal{O}(W^2LT + WLBT + NM)$} & \small{$\mathcal{O}(W^2LT + WLBT)$} \\
MILP & \small{$\mathcal{O}(WLBT + N)$} & \small{$\mathcal{O}(W^2LBT + NM)$} & \small{$0$} \\
QCP & \small{0} & \small{$\mathcal{O}(W^2LT + WLBT + NM)$} & \small{$\mathcal{O}(W^2LT + WLBT)$} \\
LP & \small{0} & \small{$\mathcal{O}(W^2LBT + NM)$} & \small{0}\\
\bottomrule
\end{tabular}
\end{table}
This section established a hierarchy of optimization approaches, ranging from exact MIQCP formulations to computationally efficient relaxations and decompositions.

Table~\ref{tab:formulations} compares our proposed formulations based on their variable and constraint complexity.
We observe that relaxing quadratic constraints, despite usually simplifying optimization, dramatically increases the formulation's size.
The practical implications of these trade-offs are explored empirically in Section~\ref{sec:results_basic}.

%% file: sections/08_results_basic.tex
\section{Core Certification Objectives and Experimental Evaluation}\label{sec:results_basic}

This section establishes the fundamental certification objectives underlying our target applications and outlines their certification via AGT. We then provide an experimental validation of the certification approach on an illustrative 2d dataset. Code to reproduce all experiments is available at \url{github.com/psosnin/AbstractGradientTraining}, with complete experimental details provided in Appendix~\ref{app:experimental}.

\subsection{Basic Certification Objectives}\label{sec:core_cert}

We introduce two point-wise certification objectives for a model's prediction on an input-label pair $(x, y)$ with respect to perturbations to the training data. Both are defined with respect to $\Theta$, the set containing all possible model parameters resulting from training on a dataset consistent with a perturbation model $\mathcal{T}$, which can be computed using AGT.

A prediction of a model $f$ is \textit{certified correct} if $f(x, \tilde{\theta}) = y$ for all $\tilde{\theta} \in \Theta$. This certificate implies that no feasible perturbation to the training data would cause the model to make an incorrect prediction on $x$. This certificate can only hold if the nominal model already predicts the correct label.

A prediction is \textit{certified stable} if $f(x, \theta) = f(x, \tilde{\theta})$ for all $\tilde{\theta} \in \Theta$. This weaker notion guarantees the prediction remains the same, but not necessarily correct, isolating the effect of training data perturbations from the model's baseline accuracy.

These certificates can be computed by deriving a sound interval bound, $\interval{\hat{y}}$, on the model's output for all $\tilde{\theta} \in \Theta$ using IBP, CROWN, or optimization formulations. If the lower bound of the logit for the predicted class $t$ exceeds the upper bounds of all other class logits (i.e., $[\hat{y}_L]_t \geq [\hat{y}_U]_{t'}$ for all $t' \neq t$) then the prediction is certified stable. If the original prediction was also correct, it is certified correct.

While the certificates discussed above apply to individual predictions, AGT's parameter-space bounds can also used to compute collective certificates over batches of data points.
Collective certificates, while providing improved tightness compared to point-wise certificates, incur a higher computational cost \citep{chen2022collective}.
Our subsequent analysis of AGT will therefore focus on point-wise certification.


\subsection{Illustrative 2d Example}
We first illustrate our approach on a binary classification task using a two-dimensional half-moons dataset with 128 training samples. We train a support vector machine (SVM) with cubic feature expansions, resulting in a 10-parameter model. We evaluate robustness in a label-flipping poisoning setting, where a single training label ($n=1$) can be flipped to the opposite class. This setup allows us to compare our certified bounds with an enumeration-based ground truth. All optimization-based bounds are solved in Gurobi on a machine with 2x AMD EPYC 9334 CPU, using a per-subproblem timeout of 3600 seconds.
\begin{figure}[ht]
    \centering
    \hfil
    \begin{subfigure}[b]{0.4\linewidth}
        \includegraphics[width=\linewidth]{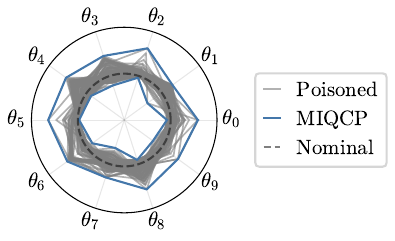}
        \caption{Element-wise parameter bounds vs empirical label flipping attacks.}
        \label{fig:halfmoons_opt_2}
    \end{subfigure}
    \hfil
    \begin{subfigure}[b]{0.4\linewidth}
        \includegraphics[width=0.95\linewidth]{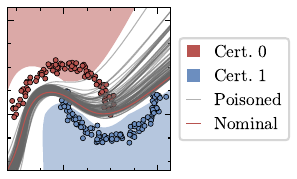}
        \caption{Decision boundaries of poisoned models vs certified regions from MIQCP.}
        \label{fig:halfmoons_opt_1}
    \end{subfigure}
    \hfil
    \vspace{1.5em}
    \hfil
    \begin{subfigure}[b]{0.7\linewidth}
        \includegraphics[width=\linewidth]{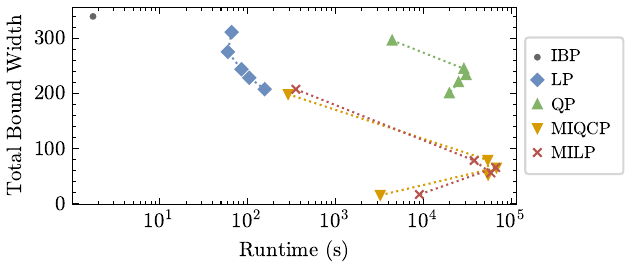}
        \caption{Runtime vs. tightness trade-off for bounding methods. Bounds using the same relaxation but different bounding strategies are grouped; Table~\ref{tab:opt_results} provides a comprehensive list of results for each formulation.}
        \label{fig:halfmoons_opt_3}
    \end{subfigure}
    \hfil
    \caption{Results of certification on the half-moons dataset and comparison with empirical label flipping attacks.}
    \label{fig:halfmoons_opt_main}
\end{figure}

\paragraph{Comparison of MIQCP with Empirical Attacks.}
We evaluate our exact MIQCP formulation against an exhaustive enumeration of all possible label flips on the dataset. Figure~\ref{fig:halfmoons_opt_2} illustrates the element-wise parameter bounds along with the parameters obtained under label flipping. Axes such are scaled for easy visual comparison.
We observe that for each parameter, the bound is tight, i.e., there is at least one label flip that causes the parameter at that index to attain its lower / upper bound.

In Figure~\ref{fig:halfmoons_opt_1}, we plot the decision boundaries of all models trained with a single label flip.
Given the small size of this example dataset, a single label flip can notably perturb the model's decision boundary.
The red and blue regions visualize where point-wise robustness certificates hold, based on the final certified parameter interval obtained from the exact MIQCP bounds. We observe only a small gap between the certified regions and the true poisoning vulnerabilities.
This gap is attributed to the independence assumption within the interval domain, which, despite the individual upper and lower parameter bounds being exact (i.e., achievable by an actual label flip), introduces some additional looseness.

\paragraph{Comparison of Bounding Methods.} 
We now compare four optimization-based formulations (LP, QP, MILP, MIQCP) and an interval-based approach (IBP). For each method, we construct interval bounds over the final model parameters by solving independent minimization and maximization problems for each parameter.

For each optimization-based method, we compare both the formulation of the full training pipeline, as well as certification via decomposition into smaller subproblems based on a \emph{rolling horizon} strategy. Specifically, we solve non-overlapping optimization subproblems spanning $w$ training iterations. Larger horizons are more precise and involve less sub-problems, but each individual solve is more costly. We also compare against IBP, which is efficient but extremely loose in this low-data setting.

Figure~\ref{fig:halfmoons_opt_3} plots the total certified bound width $\|\theta_U - \theta_L\|_1$ against runtime for each method and horizon size. We observe that all optimization-based methods outperform IBP in bound tightness, though they come with a significant runtime cost (2–5 orders of magnitude higher).
Additionally, the choice of relaxation significantly impacts both runtime and tightness. LP- and MIQCP-based bounds offer a favorable balance compared to QP and MILP formulations. MILP and MIQCP are the tightest methods, with MIQCP being the best overall in terms of both runtime and tightness for the ``Full'' formulation.

\begin{table}
\centering
\caption{Certified bound width and corresponding runtime (in seconds) across methods and horizon length (in batches). Each cell reports the certified bound width, with runtime shown in parentheses.}
\label{tab:opt_results}
\begin{tabular}{lccccc}
\toprule
\textbf{Horizon} & \textbf{IBP} & \textbf{LP} & \textbf{QP} & \textbf{MILP} & \textbf{MIQCP}\\
\midrule
1 & 339.7 (2.2) & 311.3 (66) & 296.4 (4418) & 207.5 (355) & 198.5 (290) \\
3 & -- & 275.4 (59.7) & 244.8 (28954) & 78.2 (37849) & 79.0 (54192) \\
5 & -- & 243.7 (85.5) & 233.5 (30689) & 55.8 (58724) & 52.2 (54634)\\
7 & -- & 228.5 (105) & 220.7 (25170) & 65.5 (66705) & 65.1 (67847) \\
14 (Full) & -- & 207.7 (157) & 201.9 (33157) & 16.6 (9020) & \textbf{15.3} (3240) \\
\bottomrule
\end{tabular}
\end{table}

\begin{figure}
    \centering
    \includegraphics[width=\linewidth]{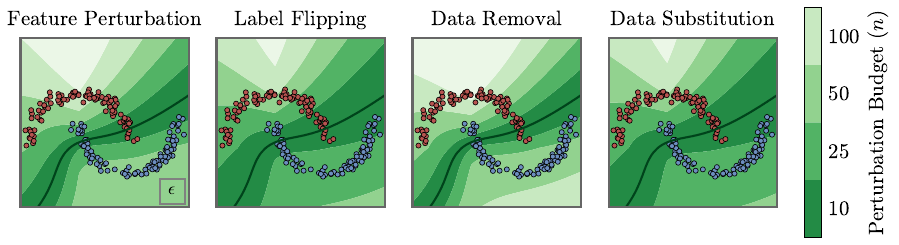}
    \caption{Regions where our certification holds for various perturbation models on a half-moons dataset. Bounds computed using interval-based AGT.}
    \label{fig:halfmoons_perturbation}
\end{figure}

\paragraph{Comparing Perturbation Models.} Finally, we evaluate the impact of various perturbation models using a larger half-moons dataset of $N=3000$ samples. We train a logistic regression model using interval-based AGT using the same cubic features as in the previous analysis. We visualize these trained models in Figure~\ref{fig:halfmoons_perturbation} for different values of perturbation budget $n$. Each colored region indicates the area for which we cannot provide certificates of prediction stability for the given perturbation model.
As expected, points far from the decision boundary of the model exhibit stronger certificates than those close to the boundary.

Of the four perturbation models considered, we observe that unlearning (data removal) yields the tightest bounds. This is due to the fact that the data removal aggregation mechanism operates only on the observed gradient bounds within each given batch. In contrast, the feature and label perturbation models must account for the worst-case adversarial modification, which necessarily admits more over-approximation. Similarly, the data substitution model, which encompasses all other perturbation models, produces strictly looser bounds.

%% file: sections/09_results_poisoning.tex
\section{Practical Case Studies: Data Poisoning}\label{sec:results_poisoning}

This section explores certified guarantees specifically in the context of adversarial data poisoning. In such attacks, an adversary manipulates a subset of the training data to degrade a model's performance. Our goal is to certify robustness against these attacks by establishing sound upper bounds on their success.

We will begin by outlining the specific certification objectives for data poisoning. We then apply them to larger-scale problems using case studies from our previous work \citep{sosnin2024certified}, focusing exclusively on interval-based Abstract Gradient Training.

\subsection{Certification Objectives and Perturbation Models}
The choice of the {perturbation model} $\mathcal{T}$ defines the adversarial capabilities that we are certifying against. For example, a \textbf{bounded adversary} that can only make bounded modifications, such as adding imperceptible triggers for backdoor attacks, is modeled by $\mathcal{T}^{n, p, \epsilon, q, \nu}_\text{bounded}$. In contrast, the more powerful \textbf{substitution model}, $\mathcal{T}^n_\text{subs.}$, represents an adversary that can arbitrarily replace entire training samples. Certification is only valid within the specific constraints of the chosen perturbation model.

We outline certified bounds for common attack goals:
\begin{itemize}
    \item \textit{Untargeted Poisoning:} These attacks aim to make the model unusable by preventing training convergence. The objective is to maximize the average loss over a dataset $\mathcal{D}_\text{cert}$:
    \begin{equation}
        J(\tilde{\theta}) = \dfrac{1}{|\mathcal{D}_\text{cert}|} \sum_{\mathcal{D}_\text{cert}} \mathcal{L}\big( f(x^{(i)}, \tilde{\theta}), y^{(i)} \big).
    \end{equation}
    An upper bound on $J(\tilde{\theta})$ provides a certificate on the attack's success.

    \item \textit{Targeted Poisoning:} The goal is to make predictions fall outside a \textbf{safe set} of outputs $S(x^{(i)}, y^{(i)})$. The objective is the fraction of data points where this occurs:
    \begin{equation}
        J(\tilde{\theta}) = \dfrac{1}{|\mathcal{D}_\text{cert}|} \sum_{\mathcal{D}_\text{cert}} \mathbbm{1}\big( f(x^{(i)}, \tilde{\theta}) \notin S(x^{(i)}, y^{(i)}) \big).
    \end{equation}
    A sound upper bound on this objective limits the attack's success rate. When the safe set is the ground-truth label, a lower bound on $1 - J(\tilde{\theta})$ is called the \textbf{certified accuracy}.

    \item \textit{Backdoor Poisoning:} These attacks introduce a latent vulnerability by assuming the adversary can also manipulate inputs at test time. The objective is to make the model produce predictions outside the safe set for a manipulated input $x^\star \in V(x^{(i)})$ (e.g., an $\ell_\infty$ ball around the input):
    \begin{equation}
        J(\tilde{\theta}) = \dfrac{1}{|\mathcal{D}_\text{cert}|} \sum_{i=1}^{k} \mathbbm{1}\big( \exists x^\star \in V(x^{(i)})\ s.t.\ f^{M}(x^\star) \notin S(x^{(i)}, y^{(i)}) \big).
    \end{equation}
    The certification for this objective must account for both training data perturbations and test-time triggers to ensure soundness.
\end{itemize}

Each of the poisoning objectives above can be bounded using the certificates described in Section~\ref{sec:core_cert}. This can be achieved through a point-wise decomposition of the batch certificates (e.g., using IBP or CROWN) or through joint consideration of $\mathcal{D}_\text{cert}$ using optimization formulations. We note that in the backdoor case, the inference-time perturbation $x^\star \in V(x^{(i)})$ must be included in any bound propagation or optimization formulation to ensure soundness with respect to trigger manipulations.

\subsection{UCI Regression}

\begin{figure}
    \centering
    \includegraphics[width=\linewidth]{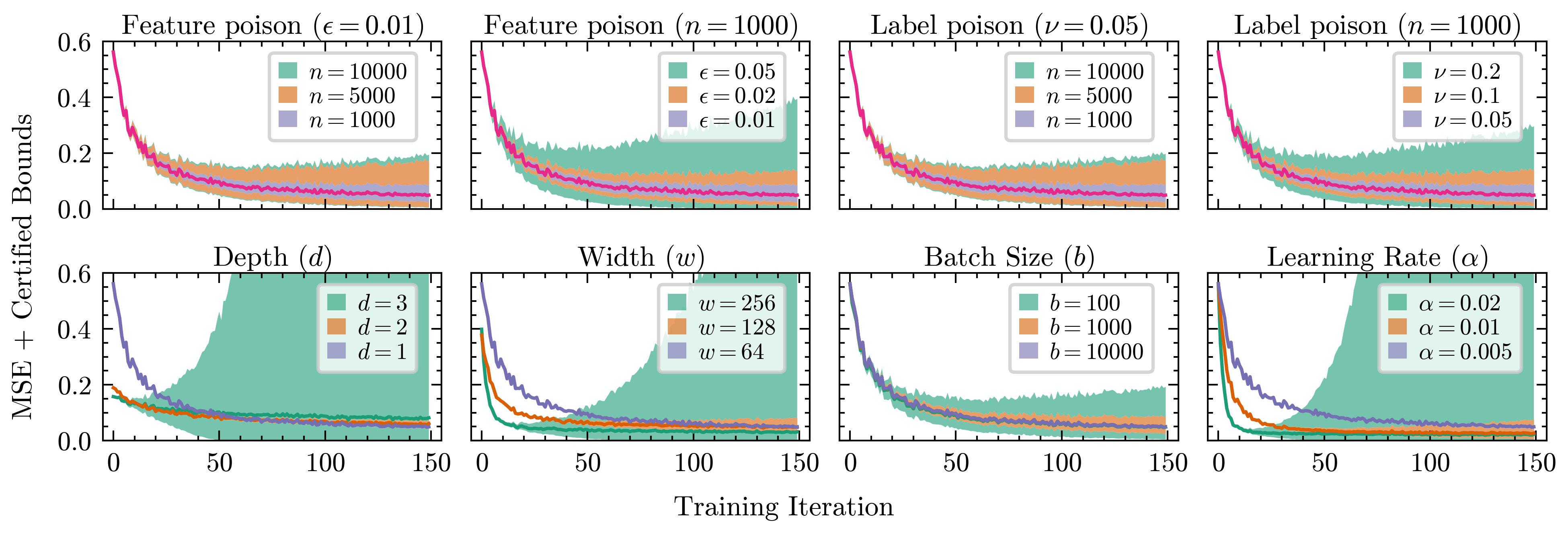}
    \caption{Mean squared error bounds on the UCI-houseelectric dataset for a bounded adversary.
    Top: Effect of adversary strength.
    Bottom: Effect of model/training hyperparameters (with $n=100, \epsilon=0.01, \nu=0$).
    Where not stated, $p=q=\infty, d=1, w=50, b=10000, \alpha=0.02$.}
    \label{fig:uci_electric_poisoning_1}
\end{figure}

We begin by examining a straightforward regression model for the household electric power consumption dataset (`houseelectric') from the UCI repository~\citep{misc_individual_household_electric_power_consumption_235}.
We train fully connected neural networks using interval-based AGT and track the evolution of our certified bounds on the model's mean squared error throughout training, as shown in Figure \ref{fig:uci_electric_poisoning_1}.
Specifically, the colored regions indicate certified upper and lower bounds on the model's mean squared error under the given training configuration and poisoning adversary.

Figure \ref{fig:uci_electric_poisoning_1} (top) illustrates the evolution of our MSE certificates for a neural network with one hidden layer of 50 neurons under various poisoning attack configurations. As expected, increasing any of the parameters $n$, $\epsilon$, or $\nu$ leads to looser certified performance bounds.
We note that we are able to obtain non-vacuous bounds for even $n=10000$, which corresponds to the entire dataset being potentially poisoned.

Figure \ref{fig:uci_electric_poisoning_1} (bottom) presents the progression of our bounds during training under a fixed poisoning attack and various hyperparameter settings.
Generally, we find that increasing the number of model parameters (increasing either in width or depth) results in weaker guarantees.
Larger batch sizes yield tighter bounds, since the fixed number of poisoned samples $n$ is distributed across more training data, reducing their relative impact.
Increasing the learning rate speeds up convergence but also causes the certified bounds to degrade more rapidly.

\subsection{Comparison with Baselines (MNIST Digit Classification)}

We evaluate our method in the multi-class setting of MNIST digit classification, focusing specifically on label-flipping attacks. Unlike in binary classification or regression, interval-based AGT struggles to produce tight guarantees in multi-class problems (see Section~\ref{sec:ibp}). To address this, we apply PCA to project the input data into a 32-dimensional feature space. Since we assume the features are clean and only the labels may be corrupted, this dimensionality reduction is certifiably sound. Similar preprocessing steps are also common in prior works on certified label poisoning, such as SS-DPA \citep{levine2020deep}.

\begin{wrapfigure}{r}{0.55\textwidth}
    \centering
    \includegraphics[width=0.95\linewidth]{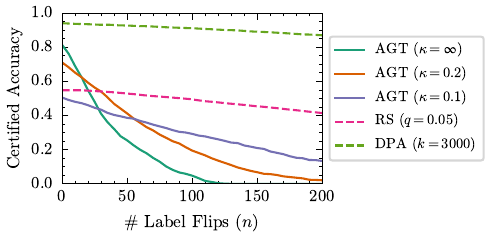}
    \vspace{-1.5em}
    \caption{Certified accuracy on the MNIST dataset under label-flipping. Curves for DPA and RS reproduced from Figures 1 of \cite{levine2020deep} and \cite{rosenfeld2020certified}, respectively.}
    \vspace{-1.5em}
    \label{fig:mnist}
\end{wrapfigure}
Figure~\ref{fig:mnist} compares the certified accuracy of our method to two existing approaches: DPA \citep{levine2020deep} and randomized smoothing (RS) \citep{rosenfeld2020certified}. While these baselines serve similar goals, they differ significantly in scope and efficiency. RS applies only to linear models and  label-flipping attacks, whereas our approach generalizes to a wider range of models and poisoning capabilities. DPA is more flexible in the types of adversaries it handles, but at the cost of substantially higher computational overhead and major changes to the training pipeline.

To illustrate this overhead, the DPA results in Figure~\ref{fig:mnist} (reproduced from \cite{levine2020deep}) use an ensemble of 3000 models. This alone introduces a 3000$\times$ increase in memory cost compared to standard training. In addition, each ensemble member is larger than our base model and trained under a less limited setup. The total runtime for generating certified guarantees with DPA is approximately 20 hours (0.4 minutes per model), whereas AGT computes certified bounds in just a few seconds.
In terms of certified performance, AGT can outperform RS when the number of flipped labels is small. However, across all tested scenarios, DPA consistently provides stronger guarantees than AGT.

\subsection{MedMNIST Image Classification}
\label{sec:poisoning_medmnist}
\begin{figure*}
    \centering
    \includegraphics[width=\linewidth]{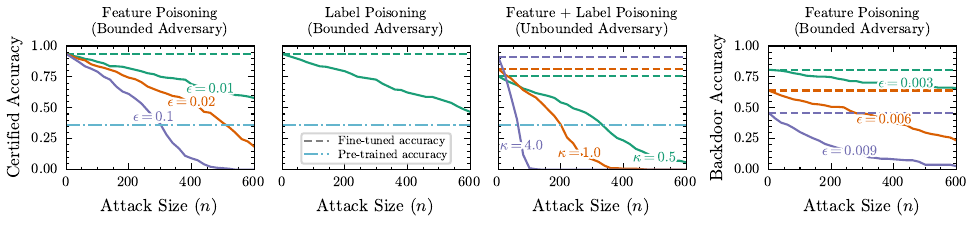}
    \caption{Certified accuracy (left) and backdoor accuracy (right) for a binary classifier fine-tuned on the Drusen class of OCTMNIST for an attack size up to 10\% poisoned data per batch ($b=6000, p=\infty, q=0, \nu=1$). 
    Dashed lines show the nominal accuracy of each fine-tuned model.}
    \label{fig:medmnist}
\end{figure*}

We next explore a fine-tuning scenario using the retinal OCT dataset (OCTMNIST)~\citep{medmnist}, which includes four classes: one normal and three abnormal. We focus on the binary classification task of distinguishing normal from abnormal samples.

Our experiments use the “small” architecture from \cite{gowal2018effectiveness}, consisting of two convolutional layers (with widths 16 and 32) followed by a fully connected layer with 100 units. The fine-tuning procedure proceeds as follows: we pre-train the model using only three of the four classes, excluding the rarest class (Drusen), via the robust training method from \cite{wicker2022robust}.
The robust pre-training ensures the base model is resilient to feature perturbations during subsequent fine-tuning.
We then simulate an attack scenario in which Drusen samples are potentially poisoned and introduced during fine-tuning as an additional abnormal class. During this phase, we train only the final dense layer using mini-batches containing a mix of 50\% Drusen samples (i.e., 3000 out of 6000 samples per batch).

Figure~\ref{fig:medmnist} presents how certified accuracy bounds degrade as the strength of the poisoning attack increases. In the case of feature-only poisoning, we find that attacks with $\epsilon > 0.02$ on roughly $n \approx 500$ samples are sufficient to produce certified bounds that fall below the accuracy of the original pre-trained model. When allowing both feature and label poisoning, the bounds degrade more rapidly, as expected. Increasing the clipping parameter $\kappa$ can improve certified accuracy in this setting, at the expense of nominal model performance (Figure~\ref{fig:medmnist}, center right).
Under label-only poisoning, the model proves more robust: $n \geq 600$ mislabeled samples are required for the certified accuracy bound to drop to that of the original pre-trained model, corresponding to 20\% of Drusen samples per batch being mislabeled.

Finally, we consider a backdoor attack scenario in which the adversary’s strength at training and inference time is matched. Even without poisoning, the model is vulnerable to small adversarial perturbations at inference time: an $\epsilon$ as low as 0.009 is sufficient to reduce certified accuracy on backdoored inputs to below 50\%. As the adversary’s strength increases, the certified accuracy continues to decline.

\begin{figure}
    \centering
\includegraphics[width=0.28\textwidth, valign=m]{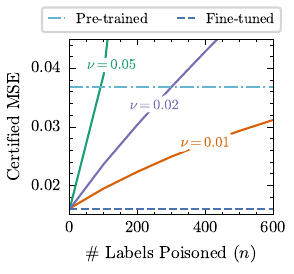}
\includegraphics[width=0.7\textwidth, valign=m, raise=0.4em]{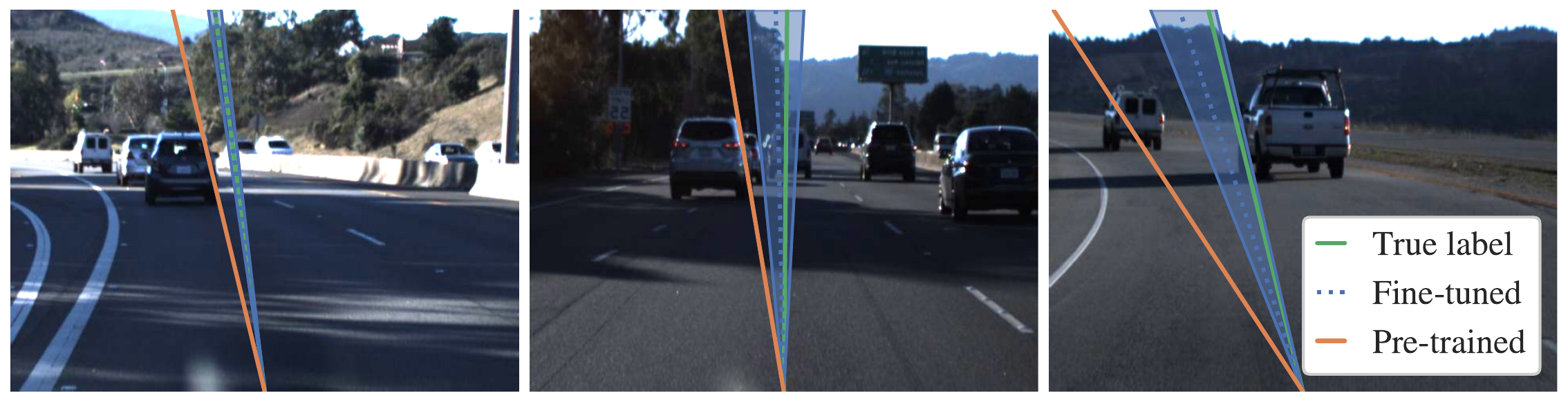}
    \caption{Left: Fine-tuning PilotNet on unseen data with a bounded label poisoning attack ($q=\infty$). Right: We plot the steering angle prediction before and after fine-tuning ($n=300, q=\infty, \nu=0.01$). The angle of the lines indicates the predicted steering angle.}
    \label{fig:pilotnet}
\end{figure}

\subsection{Data Poisoning in Self-Driving Applications}

Finally, we apply our method to a steering angle prediction task for autonomous driving, where the goal is to predict the vehicle’s steering angle from an input image~\citep{bojarski2016end}. The model architecture consists of convolutional layers with 24, 36, 48, and 64 filters, followed by fully connected layers with 100, 50, and 10 units.

Our fine-tuning setup mirrors the previous experiment. The model is first pre-trained on videos 2 through 6 of the Udacity self-driving car dataset\footnote{\url{github.com/udacity/self-driving-car/tree/master}}.
We then fine-tune only the dense layers using data from video 1, which features challenging lighting conditions, while allowing for the possibility of label poisoning during this phase.

Figure~\ref{fig:pilotnet} presents certified bounds on mean squared error (MSE) for the video 1 data and illustrates how these bounds impact predicted steering angles. As in previous experiments, fine-tuning improves nominal performance on the new data. However, the certified MSE bounds degrade as the number of potentially poisoned samples increases (Figure~\ref{fig:pilotnet}, left), with the rate of degradation highly sensitive to the poisoning strength parameter $\nu$.

%% file: sections/10_results_privacy.tex
\section{Applications in Unlearning and Differential Privacy}\label{sec:results_privacy}

\subsection{Machine Unlearning}

Certification with respect to the removal of data points is of particular interest in the field of machine unlearning.
While our method does not fit precisely within existing machine unlearning settings, our bounds deterministically bound the distance between a trained model and an unlearned model. By employing our parameter space bounds, we can characterize aspects of model behavior under unlearning, such as prediction stability.
State-of-the-art approaches in certified machine unlearning, such as SISA \citep{bourtoule2021machine}, typically operate by training an ensemble of classifiers on disjoint sets of user data.
Upon receiving an unlearning request, the affected model is taken offline to undergo partial re-training.

A potential application of our bounds is to allow certification that individual model predictions would remain unchanged under re-training with up to $n$ data points removed.
Specifically, we can achieve this by first training a model using AGT under the perturbation model $\mathcal{T}_\text{removal}^n$; at inference time, we can then compute certificates of prediction stability with respect to the resulting parameter-space domain.
This capability has the following practical implications: if such certificates are provided for all queries to the model, the system may potentially remain online with no risk to user data privacy, provided that the number of unlearning requests remains below $n$.

\subsection{Differentially Private Prediction}

Finally, we outline the usage of our parameter-space bounds in differentially private prediction. Here we provide an overview of a procedure to bound the \textit{smooth sensitivity} of model predictions; for further details on this quantity and a comprehensive privacy analysis of our approach, we refer the reader to our previous work \citep{wicker2024certification}.

\subsubsection{Prediction Sensitivity}

In differential privacy, the \textit{sensitivity} of a prediction is a measure of how much the output of a function, such as a model's prediction, can change when a single individual's data is added to or removed from the dataset. The sensitivity can be used to calibrate additive noise that can ensure that a model's output satisfies Definition~\ref{def:approxdp} for a given $(\epsilon, \delta)$.

The \textit{global sensitivity} represents the largest possible change in a function's output when comparing any two datasets that differ by a single data point. For instance, in binary classification, the maximum possible change in the classifier's output is 1.

However, for many predictions, global sensitivity significantly overestimates the actual change in the model's output that results from a single data change. This overestimation leads to adding an excessive amount of noise to the prediction to preserve privacy, which can degrade the model's accuracy. This has led to the development of tighter, data-dependent notions of sensitivity, namely the \textit{smooth sensitivity}:
\begin{definition}[Smooth Sensitivity, \cite{nissim2007smooth}]\label{def:smoothsens}
    The $\beta$-smooth\footnote{Here $\beta$ is a parameter that takes any value $\leq \epsilon / 6$.} sensitivity of the model prediction $f(x, \theta)$ trained on a dataset $\mathcal{D}$ via training mechanism $M$ is
    \begin{equation}
        \operatorname{SS}^\beta(f(x, \cdot), \mathcal{D}) = \max\limits_{n \in \mathbb{N}^+} e^{-\beta n} A^n(f(x, \cdot), \mathcal{D})
    \end{equation}
    where $A^n(f, \mathcal{D})$ is the \textit{maximum local sensitivity}, defined as
    \begin{equation}
    A^n(f(x, \cdot), \mathcal{D}) = \max\limits_{\tilde{\mathcal{D}} \in \mathcal{T}^n_\text{subs.}(\mathcal{D}) } \operatorname{LS}(f(x, \cdot), \tilde{\mathcal{D}}).
    \end{equation}
    Here, $\operatorname{LS}(f(x, \cdot), \mathcal{D})$ is the \textit{local sensitivity}, given by
    \begin{equation}
        \operatorname{LS}(f(x, \cdot), \mathcal{D}) = \max\limits_{\tilde{\mathcal{D}} \in \mathcal{T}^1_\text{subs.}(\mathcal{D})} \|f(x, \theta) - f(x, \tilde{\theta})\|_1 \text{ s.t. } \theta = M(f, \mathcal{D}), \tilde{\theta} = M(f, \tilde{\mathcal{D}}).
    \end{equation}
\end{definition}
Our previous work established the following result that constitutes a sound upper bound on the smooth sensitivity in terms of point-wise certificates of prediction stability.
\begin{theorem}[Certified Upper Bound on Smooth Sensitivity, \cite{wicker2024certification}]\label{thrm:smooth_sens_bound}
    Let $f(x, \theta)$ denote the prediction of a binary classifier with parameters $\theta$ trained on a dataset $\mathcal{D}$ at a point $x$. 
    If $f(x, \theta)$ is stable at a distance of $n'$ with respect to the perturbation model $\mathcal{T}_\text{subs.}^{n'}$, then the following provides an upper bound on the $\beta$-smooth sensitivity:
\begin{equation}
        \operatorname{SS}^\beta(f(x, \cdot), D) = \max\limits_{n \in \mathbb{N}^+} e^{-\beta n} A^n(f(x, \cdot), D) \leq e^{-\beta n'}
\end{equation}
\end{theorem}
To use the above, we first compute valid parameter space bounds for a fixed set of $n$ values. Then, for a given query $x$, we attempt to compute certificates of prediction stability for the largest possible $n$ within our set. Then Theorem~\ref{thrm:smooth_sens_bound} can be applied to upper-bound the smooth sensitivity.

\begin{figure}
    \centering
    \includegraphics[width=\linewidth]{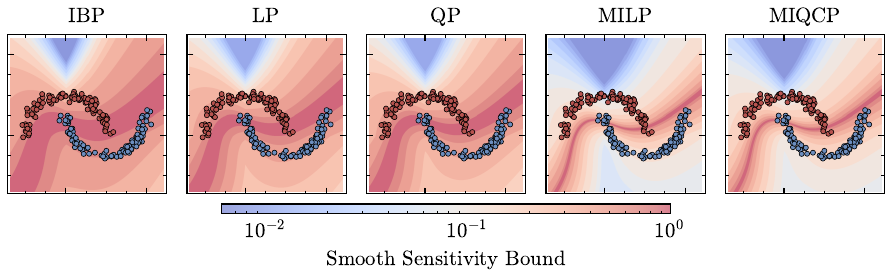}
    \caption{Visualization of smooth sensitivity bounds obtained using various AGT bounding strategies for privacy parameters $\epsilon=2.0, \delta=0$. A horizon length of 1 is used for all optimization-based methods.}
    \label{fig:halfmoons_smooth_sens}
\end{figure}

\paragraph{Experimental Validation of Smooth Sensitivity Bounds.} Figure~\ref{fig:halfmoons_smooth_sens} depicts the upper bound on smooth sensitivity of a model trained on the halfmoons dataset ($N=3000$). For each bounding method, parameter-space bounds are computed for $n=1, \dots, 15$ during model training. These bounds are then used to upper-bound the smooth sensitivity according to Theorem~\ref{thrm:smooth_sens_bound}.
Noting that the global sensitivity for binary classification takes a value of 1, we observe that along the decision boundary, the smooth sensitivity bound cannot improve over the global bound. However, for predictions further from the decision boundary the smooth sensitivity can have smooth sensitivity bounds several orders of magnitude lower than the global sensitivity, with tighter bounding achieving the tightest bounds. As we will demonstrate below, having tighter bounds on the smooth sensitivity leads to improved downstream performance in the private prediction setting.  

\subsubsection{Private Prediction using Smooth Sensitivity}

In order to ensure that the the prediction of a model satisfies differential privacy, it is necessary to add noise to the model outputs before releasing them. In binary classification, for example, we can privatize predictions using the following response mechanism:  
\begin{equation}
    R(x) =  
    \begin{cases}  
    1, & \text{if } f(x, \theta) + z > 0.5, \\  
    0, & \text{otherwise},  
    \end{cases}  
\end{equation}
where $z$ is a noise term drawn from a distribution chosen to satisfy Definition~\ref{def:approxdp}.  

A common approach is the \textit{Laplace mechanism}, which adds noise from $\operatorname{Lap}(1/\epsilon)$, ensuring $(\epsilon, 0)$-differential privacy. However, this approach relies on global sensitivity, leading to excessive noise in many cases.  

To improve privacy-utility tradeoffs, \citet{nissim2007smooth} showed that drawing  
\begin{equation}
    z \sim \operatorname{Cauchy}\left(\frac{6 \operatorname{SS}^\beta(f(x, \cdot), \mathcal{D})}{\epsilon} \right)
\end{equation}  
ensures $(\epsilon, 0)$-differential privacy for $\beta \leq \epsilon / 6$. Moreover, any upper bound on $\operatorname{SS}^\beta(f(x, \cdot), \mathcal{D})$ can be used to calibrate noise.  

We use AGT to compute a certified upper bound on smooth sensitivity, which we then use to adjust the noise level. This significantly reduces the magnitude of noise required to meet Definition~\ref{def:approxdp}, thereby enhancing the privacy-utility tradeoff in private prediction.  

\paragraph{Experimental Results in Private Prediction.} Here we reproduce the results from \cite{wicker2024certification} that studies the effectiveness of our proposed mechanism across three tasks: (i) Blobs, training a logistic regression model on a synthetic dataset of isotropic Gaussian clusters, (ii) Medical imaging: fine-tuning a model on retinal OCT images analogous to Section~\ref{sec:poisoning_medmnist}, and (iii) Sentiment Classification, training a neural network for sentiment analysis on the IMDB movie reviews dataset, using GPT-2 embeddings as input features.

\begin{figure*}
    \centering
    \includegraphics[width=\linewidth]{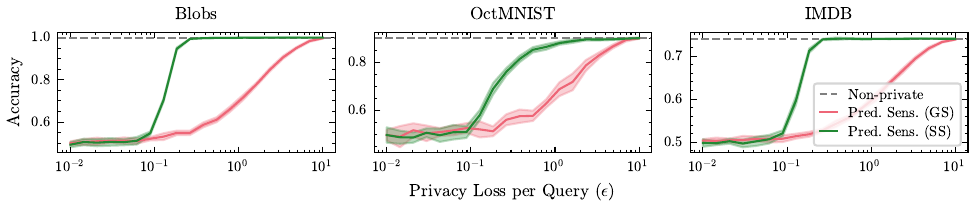}
    \vspace{-1em}
    \caption{Private prediction accuracy for a single model using global sensitivity vs our smooth sensitivity bounds derived using interval-based AGT.}
    \label{fig:single_model}
    \vspace{-1em}
\end{figure*}

Figure~\ref{fig:single_model} compares the performance of the prediction sensitivity mechanism using global sensitivity versus smooth sensitivity bounds obtained using interval-based AGT.
Our method maintains near noise-free accuracy at privacy budgets ($\epsilon$) up to an order of magnitude smaller than those required when using global sensitivity. This improvement is especially pronounced in settings where the data are well-separated (as in the "Blobs" task) or when the training dataset is large (as in the "IMDB" task).

%% file: sections/11_conclusions.tex
\section{Conclusions}
In this work, we introduced a mathematical framework for computing rigorous parameter-space bounds on the impact of training data perturbations, including data poisoning, data removal, and data substitution. Our framework defines general perturbation models that capture various forms of training data manipulation and systematically propagates their effects throughout the entire training process.

We presented two instantiations of our framework: an efficient but conservative over-approximation using interval bound propagation and an exact analysis via mixed-integer programming.
Through our experiments, we analyzed the trade-offs between these approaches, highlighting the ongoing need for tighter yet computationally efficient bounding techniques. Finally, we demonstrated the practical effectiveness of our method on real-world applications, including autonomous driving and medical image classification. This work lays a foundation for future certification-based techniques aimed at enhancing the robustness and privacy of machine learning models.

%% file: sections/12_appendix_ibp.tex
\section{Interval Propagation Through Common Loss Functions}\label{app:ibp_loss_fn}

\textit{Cross Entropy Loss.} The cross-entropy loss, defined as
\begin{equation}
    \mathcal{L}\left(\hat{y}, y\right) = - \sum_{i} y^t_i \log p_i,
\end{equation}
with gradient
\begin{equation}
    \frac{\partial\mathcal{L}\left(\hat{y}, y\right)}{\partial \hat{y}} = p - y^t
\end{equation}
measures the distance between the predicted probability distribution and the true labels.
Here, $p=\operatorname{softmax}(\hat{y})$ are the predicted class probabilities and $y^t$ is the one-hot encoding of the true class $y$.

To compute a sound enclosure on the loss gradient, we first derive interval bounds on the output probabilities $p_i$, which are obtained by passing the model output $\hat{y}$ through the softmax function, given by
\begin{equation}
    p_i = \frac{1}{1 + \sum_{j\neq i} \exp{\left(\hat{y}_j - \hat{y}_i\right)}}.
\end{equation}
Sound bounds on the output probabilities are obtained by reasoning over the output interval $\boldsymbol{\hat{y}}$:
\begin{equation}
 \boldsymbol{p_i} = \left[\frac{1}{1 + \sum_{j\neq i} \exp{\left(\hat{y}^U_j - \hat{y}^L_i\right)}}, \frac{1}{1 + \sum_{j\neq i} \exp{\left(\hat{y}^L_j - \hat{y}^U_i\right)}}     \right]   
\end{equation}
Now, taking $\boldsymbol{y^t}$ to be the interval over the one-hot encoding of the (potentially poisoned) true label, the interval over the loss gradient is given by
\begin{equation}
    \boldsymbol{\frac{\partial\mathcal{L}\left(\hat{y}, y\right)}{\partial \hat{y}}} = \boldsymbol{p} \ominus \boldsymbol{y^t}
\end{equation}

\textit{Hinge Loss.}
The hinge loss, commonly used for training support vector machines (SVMs), is defined as $\mathcal{L}(\hat{y}, y) = \max(0, 1 - y \hat{y})$  
The gradient of the hinge loss with respect to the predicted output $\hat{y}$ is given by:  
\begin{equation}
\frac{\partial \mathcal{L}}{\partial \hat{y}} =
\begin{cases}
    -y, & \text{if } 1 - y \hat{y} > 0\\
    0, & \text{otherwise}.
\end{cases}
\end{equation}
To compute an interval enclosure for this gradient, we first note that both $\hat{y}$ and $y$ may be perturbed under the given threat model. We define the interval representations:
\begin{itemize}
    \item $\boldsymbol{\hat{y}} = [\hat{y}^L, \hat{y}^U]$, obtained from forward pass bounds. 
    \item $\boldsymbol{y} = [y, y]$ without label perturbation, $\boldsymbol{y} = [0, 1]$ with label perturbation.  
\end{itemize}

%% file: sections/13_appendix_crown.tex
\section{Improved Forward Pass Bounds using an Extended CROWN Algorithm}\label{app:crown}

In Section~\ref{sec:per-sample-propagation}, we extended the interval-based certification framework of \cite{gowal2018effectiveness} for computing forward pass bounds to consider intervals over the parameters of the network.
A natural progression is to explore the applicability to other certification algorithms to this setting.
To address this, we present a new, explicit extension of the CROWN algorithm \citep{zhang2018efficient} that handles interval-bounded weights and biases.
Here we present only a summary of the proposed method. For a more comprehensive overview along with proofs of all results we refer the reader to \cite{sosnin2024certified}.

The standard CROWN algorithm bounds the outputs of a neural network's $m$-th layer by back-propagating linear bounds through each intermediate activation function to the input layer.
In the case of interval parameters, the sign of a particular weight may be ambiguous (when the interval spans zero), making it impossible to determine which linear bound (upper or lower) to back-propagate. 

Our approach below generalizes the CROWN algorithm to the case of interval parameters, where the weights and biases involved in each linear relaxation are themselves represented as intervals.
We note that linear bound propagation with interval parameters has been studied previously in the context of floating-point sound certification \citep{singh2019}.

Following the notation of the original CROWN algorithm \citep{zhang2018efficient}, we present our result for an extended CROWN algorithm that bounds outputs for neural networks with interval-bounded parameters:

\begin{proposition}[Explicit output bounds of neural network $f$ with interval parameters, \cite{sosnin2024certified}]\label{prop:crown}
Given an $m$-layer neural network function $f: \mathbb{R}^{n_\text{in}} \rightarrow \mathbb{R}^{n_\text{out}}$ whose unknown parameters lie in the intervals ${b}^{(k)} \in \boldsymbol{b^{(k)}}$ and $W^{(k)} \in \boldsymbol{W^{(k)}}$ for $k=1, \dots, m$, there exist two explicit functions
\begin{align}
    f_j^L\left({x}, {\Omega}^{(0:m)}, {\Theta}^{(1:m)}, {b}^{(1:m)}\right)
    ={\Omega}_{j,:}^{(0)} {x}+\sum_{k=1}^m {\Omega}_{j,:}^{(k)}\left({b}^{(k)} +{\Theta}_{:, j}^{(k)}\right)\\
    f_j^U\left({x}, {\Lambda}^{(0:m)}, {\Delta}^{(1:m)}, {b}^{(1:m)}\right)=
    {\Lambda}_{j,:}^{(0)} {x}+\sum_{k=1}^m {\Lambda}_{j,:}^{(k)}\left({b}^{(k)} +{\Delta}_{:, j}^{(k)}\right)
\end{align}
such that $\forall x \in \boldsymbol{x}$
\begin{align*}
    f_j(x) \geq \min \left\{f_j^L\left({x}, {\Omega}^{(0:m)}, {\Theta}^{(1:m)}, {b}^{(1:m)}\right) \mid {\Omega}^{(k)} \in \boldsymbol{\Omega^{(k)}}, {b}^{k} \in {\boldsymbol{b^{(k)}}}\right\}  \\
    f_j({x}) \leq \max \left\{f_j^U\left({x}, {\Lambda}^{(0:m)}, {\Delta}^{(1:m)}, {b}^{(1:m)}\right) \mid {\Lambda}^{(k)} \in {\boldsymbol{\Lambda^{(k)}}}, {b}^{k} \in {\boldsymbol{b^{(k)}}} \right\},
\end{align*}
where $\boldsymbol{x}$ is a closed input domain and ${\Lambda}^{(0:m)}, {\Delta}^{(1:m)}, {\Omega}^{(0:m)}, {\Theta}^{(1:m)}$ are the equivalent weights and biases of the upper and lower linear bounds, respectively.
The bias terms ${\Delta}^{(1:m)}, {\Theta}^{(1:m)}$ are explicitly computed based on the linear relaxation of the activation functions. 
The weights ${\Lambda}^{(0:m)}, {\Omega}^{(0:m)}$ lie in intervals $\boldsymbol{\Lambda^{(0:m)}}, \boldsymbol{\Omega^{(0:m)}}$, which are computed in an analogous way to standard (non-interval) CROWN.
\end{proposition}

\paragraph{Computing Equivalent Weights and Biases.}
Our instantiation of the CROWN algorithm in Proposition~\ref{prop:crown} relies on the computation of the equivalent bias terms ${\Delta}^{(1:m)}, {\Theta}^{(1:m)}$ and interval enclosures over the equivalent weights ${\Omega}^{(0:m)}, {\Lambda}^{(0:m)}$.
This proceeds similarly to the standard CROWN algorithm, but now accounts for intervals over the parameters $b^{(1:m)}, W^{(1:m)}$ of the network.

\paragraph{Closed Form Output Bounds.} Given the two functions
$f_j^L(\cdot)$, $f_j^U(\cdot)$ as defined above and intervals over all the relevant variables, we can compute the following closed-form global bounds:
\begin{align*}
    \gamma^L_j = \min\left\{
    {\boldsymbol{\Omega^{(0)}_{{j,:}}}} \otimes \boldsymbol{x} \oplus
    \sum_{k=1}^m {\boldsymbol{\Omega^{(k)}_{j,:}}} \otimes \left[{\boldsymbol{b^{(k)}}}
    \oplus
    {{\Theta}}^{(k)}_{:, j}\right]\right\}\\
    \gamma^U_j = \max\left\{
    {\mathbf{\Lambda}}^{(0)}_{{j,:}} \otimes \boldsymbol{x} \oplus
    \sum_{k=1}^m {\boldsymbol{\Lambda^{(k)}_{j,:}}} \otimes \left[{\boldsymbol{b^{(k)}}}
    \oplus
    {{\Delta}}^{(k)}_{:, j}\right]\right\}
\end{align*}
where $\min/\max$ are performed element-wise and return the lower / upper bounds of each interval enclosure. 
Then, we have $\gamma^L_j \leq f_j({x}) \leq \gamma^U_j$ for all ${x} \in \boldsymbol{x}$, ${b}^{(k)} \in \boldsymbol{b^{{(k)}}}$ and ${W}^{(k)} \in {\boldsymbol{W^{{(k)}}}}$, which suffices to bound the output of the neural network. These intermediate bounds can then be used to further bound the gradient of the network as in Section~\ref{sec:ibp}.

\paragraph{Bounding the Backward Pass using CROWN.}
The CROWN algorithm can be applied to any composition of functions that can be upper- and lower-bounded by linear equations.
Therefore, it is possible to consider both the forward and backward passes in a single combined CROWN pass.
However, linear bounds on the gradient of the loss function tend to be relatively loose, e.g., linear bounds on the softmax function may be orders-of-magnitude looser than constant $[0,1]$ bounds \citep{wei2023convex}.
As a result, we found that the tightest bounds are obtained by using either IBP / CROWN to bound the forward pass and IBP to bound the backward pass.

%% file: sections/14_appendix_experimental.tex
\section{Experimental Details}\label{app:experimental}

This section details the experimental set-up used for the experiments detailed in Sections~\ref{sec:results_basic}-\ref{sec:results_privacy}. All experiments were run on a server equipped with 2x AMD EPYC 9334 CPUs and 2x NVIDIA L40 GPUs using an implementation of Abstract Gradient Training written in Python available at \url{github.com/psosnin/AbstractGradientTraining}. All optimization-based bounds are solved in Gurobi with a limit of 16 threads per subproblem with a timeout of 3600 seconds. 

Tables~\ref{tab:datasets} and \ref{tab:hyperparams} detail the specific dataset and hyperparameter settings used in our experimental evaluation. For each experimental result, we provide the number of epochs, learning rate, learning rate decay factor ($\eta$), and batchsize.
Learning rate decay is applied using a standard learning rate schedule $\alpha_n = \alpha / (1 + \eta n)$.
Below we provide further details for specific settings not covered in the main text.

\begin{table}
\centering
\begin{tabular}{l|l|l}
Dataset           & \# Features & \# Samples  \\
\midrule
Halfmoons (small) & 9$^\dagger$           & 128        \\
Halfmoons (large) & 9$^\dagger$           & 3000       \\
Blobs  & 6$^\dagger$ & 3000 \\
UCI House-electric & 11             & 2049280\\
MNIST & 784 & 60000\\
OCT-MNIST & 784 & 97477 \\
Udacity Self-Driving & 39600 & 31573\\
IMDB & 768$^\star$ & 40000
\end{tabular}
\caption{Datasets used in experimental evaluations. $\dagger:$ feature dimension including polynomial feature expansion. $^\star:$ feature dimension post GPT-2 embedding.}
\label{tab:datasets}
\end{table}

\begin{table}
\centering
\footnotesize{
\begin{tabularx}{\textwidth}{c|ccccccc}
{Figure \& Dataset)} & {Model} & Loss & \makecell[t]{Learn.\\rate} & \makecell[t]{Batch\\Size} &\makecell[t]{\#\\Epochs} & \makecell[t]{Learn.\\rate\\decay} & \makecell[t]{Gradient\\Clipping} \\
\midrule
\midrule
\makecell{Figure~\ref{fig:halfmoons_opt_main}\\Halfmoons (small)} & Linear & Hinge & 5.0 & 64 & 7 & -- & -- \\
\midrule
\makecell{Figure~\ref{fig:halfmoons_perturbation}\\Halfmoons (large)} & Linear & BCE & 3.0 & 3000 & 10 & 0.6 & 0.5\\
\midrule
\makecell{Figure~\ref{fig:uci_electric_poisoning_1}\\UCI Houseelectric} & FC & MSE & \makecell{0.005-\\0.02} & \makecell{100-\\10000} & 1 & -- & -- \\
\midrule
\makecell{Figure~\ref{fig:mnist}\\MNIST} & PCA + Linear & BCE & 5.0 & 60000 & 3 & 1.0 & --\\
\midrule
\makecell{Figure~\ref{fig:medmnist}\\OCT-MNIST} & \makecell{Conv. Small\\ \citep{gowal2018effectiveness}} & BCE & 0.05 & 6000 & 2 & 5.0 & --\\
\midrule
\makecell{Figure~\ref{fig:pilotnet}\\Udacity Self-Driving} & \makecell{PilotNet\\ \citep{bojarski2016end}} & MSE & 0.25 & 10000 & 2 & 10.0 & --\\
\midrule
\makecell{Figure~\ref{fig:halfmoons_smooth_sens}\\Halfmoons (large)} & Linear & Hinge & 2.0 & 3000 & 9 & -- & --\\
\midrule
\makecell{Figure~\ref{fig:single_model}\\Blobs} & Linear & BCE & 1.0 & 3000 & 4 & 0.6 & 0.06 \\
\midrule
\makecell{Figure~\ref{fig:single_model}\\OCT-MNIST} & \makecell{Conv. Small\\ \citep{gowal2018effectiveness}} & BCE & 0.06 & 20000 & 4 & 0.5 & 0.9 \\
\midrule
\makecell{Figure~\ref{fig:single_model}\\IMDB} & FC & BCE & 0.2 & 100000 & 3 & 0.5 & 0.02
\end{tabularx}
}
\caption{Hyperparameter settings used in experimental evaluations.}
\label{tab:hyperparams}
\end{table}

\subsection{Additional Experimental Details: Differentially Private Prediction}

In Figure~\ref{fig:single_model}, we examine the performance of AGT in differentially private prediction for three distinct binary classification tasks. Below we provide additional experimental details describing each setting in detail.

\paragraph{Blobs Dataset.}
In Figure~\ref{fig:single_model} (left), we examine a dataset consisting of 3,000 samples drawn from an isotropic Gaussian distribution. 
The objective is to predict the distribution of origin for each sample in a supervised learning task.
We train a single-layer neural network with 128 hidden neurons using interval-based AGT for values of $n$ in $\{1, \dots, 100\}$. 
Using these parameter-space bounds, we determine the maximum stable $n^\star \in \{1, \dots, 100\}$ for each point in a grid over the domain. 
This value is used to compute a bound on the smooth sensitivity, which is further used to calibrate noise for the downstream private prediction task.

\paragraph{Retinal OCT Image Classification.}
Next, we consider another dataset with larger scale inputs: classification of medical images from the retinal OCT (OctMNIST) dataset of MEDMNIST~\citep{medmnist}. 
We consider binary classification over this dataset, where a model is tasked with predicting whether an image is normal or abnormal (the latter combines three distinct abnormal classes from the original dataset). 

The model comprises two convolutional layers of 16 and 32 filters and an ensuing 100-node dense layer, corresponding to the `small' architecture from \cite{gowal2018effectiveness}.
To demonstrate our framework, we consider a base model pre-trained on public data and then fine-tuned on new, privacy-sensitive data, corresponding to the 7754 Drusen samples (a class of abnormality omitted from initial training). 
First, we train the complete model excluding this using standard stochastic gradient descent.
We then fine-tune only the dense layer weights to recognise the new class, with a mix of 50\% Drusen samples per batch. 
We aim to ensure privacy only with respect to the fine-tuning data.

\paragraph{IMDB Movie Reviews}
Finally, we consider fine-tuning GPT-2 \citep{radford2019language} for sentiment analysis on the large-scale (40,000 samples) IMDb movie review dataset~\citep{imdb}. 
In this setup, we assume that GPT-2 was pre-trained on publicly available data, distinct from the data used for fine-tuning, which implies no privacy risk from the pre-trained embeddings themselves.
Under this assumption, we begin by encoding each movie review into a 768-dimensional vector using GPT-2’s embeddings.
We then train a fully connected neural network consisting of $1\times100$ nodes, to perform binary sentiment classification (positive vs negative reviews).

%% file: sections/15_appendix_proofs.tex
\section{Proofs}

\subsection{Proof of Theorem~\ref{thm:paramcerts} (Parameter Space Equivalence)}

\begin{proof}
    Without loss of generality, take the objective function we wish to optimize to be denoted simply by $J$. By definition, there exists a parameter, $\theta^\circ = M(\mathcal{\tilde{D}})$ resulting from a particular dataset $\tilde{\mathcal{D}} \in \mathcal{T}(\mathcal{D})$ such that $\theta^\circ$ provides a (potentially non-unique) optimal solution to the optimization problem we wish to bound, i.e., the left hand side of the inequality. Given a valid parameter space domain $\Theta$ satisfying \eqref{eq:parambounds}, we have that necessarily, $\theta^\circ \in \Theta$. Therefore, the result of optimizing over $\Theta$ can provide at a minimum the bound realized by $\theta^\circ$; however, due to approximation, this bound might not be tight, so optimizing over $\Theta$  provides an upper-bound, thus proving the inequalities above.
\end{proof}

\subsection{Proof of Theorem~\ref{thm:aggregation_unlearning} (Sound Aggregation for Data Removal)}

The nominal descent direction for a parameter $\theta$ is the averaged gradient over a training batch $\mathcal{B}$, defined as
$$
\Delta \theta = \frac{1}{b} \sum_{i=1}^b \delta^{(i)}
$$
where each gradient term is given by $\delta^{(i)} = \nabla_\theta \mathcal{L} \left( f^\theta \left( x^{(i)} \right), y^{(i)} \right)$.
Our goal is to upper bound this descent direction when up to $n$ points from $\mathcal{B}$ are removed.
We wish to compute this bound to additionally be sound with respect to any $\theta \in [\theta_L, \theta_U]$. We will begin by bounding the descent direction for a fixed $\theta$, then generalize to all $\theta \in [\theta_L, \theta_U]$. We present only the upper bounds here, though corresponding results for the lower bound can be shown by reversing the inequalities and replacing $\operatorname{SEMax}$ with $\operatorname{SEMin}$.

\textbf{Bounding the descent direction for a fixed, scalar $\theta$.} Consider the effect of removing up to $n$ data points from batch $\mathcal{B}$. Without loss of generality, assume the gradient terms are sorted in descending order, i.e., $\delta^{(1)} \geq \delta^{(2)} \geq \dots \geq \delta^{(b)}$. Then, the average gradient over all points can be bounded above by the average over the largest $b - n$ terms:
$$
\Delta \theta = \frac{1}{b} \sum_{i=1}^b \delta^{(i)} \leq \frac{1}{b - n} \sum_{i=1}^{b - n} \delta^{(i)}
$$
This bound corresponds to removing the $n$ points with the smallest gradients.

\textbf{Bounding the effect of a variable parameter interval.} We extend this bound to any $\theta \in [\theta_L, \theta_U]$. Assume the existence of upper bounds $\delta^{(i)}_U$ on the gradients for each data point over the interval, such that
$$
\delta^{(i)}_U \geq  \nabla_{\theta'} \mathcal{L} \left( f^{\theta'}(x^{(i)}), y^{(i)} \right) \quad \forall \, \theta' \in [\theta_L, \theta_U].
$$
Then, using these upper bounds, we further bound $\Delta \theta$ as
$$
\Delta \theta \leq \frac{1}{b - n} \left(\sum_{i=1}^b \delta^{(i)}_U \right)
$$
where, as before, we assume $\delta^{(i)}_U$ are indexed in descending order.

\textbf{Extending to the multi-dimensional case.} To generalize to the multi-dimensional case, we apply the above bound component-wise. Since gradients are not necessarily ordered for each parameter component, we introduce the $\operatorname{SEMax}_n$ operator, which selects and sums the largest $n$ terms at each index. This yields the following bound on the descent direction:
$$
\Delta \theta \leq \frac{1}{b-n} \left( \underset{b-n}{\operatorname{SEMax}} \left\{ \delta^{(i)}_U \right\}_{i=1}^b\right)
$$
which holds for any $\theta \in [\theta_L, \theta_U]$ and up to $n$ removed points.
\qed

We have established the upper bound on the descent direction. The corresponding lower bound can be derived by reversing the inequalities and substituting $\operatorname{SEMax}$ with the analagous minimization operator, $\operatorname{SEMin}$.

\subsection{Proof of Theorem~\ref{thm:aggregation_privacy} (Sound Aggregation for Data Substitution)}

The nominal clipped descent direction for a parameter $\theta$ is the averaged, clipped gradient over a training batch $\mathcal{B}$, defined as
$$
\Delta \theta = \frac{1}{b} \sum_{i=1}^b \operatorname{Clip}_\kappa \left[ \delta^{(i)} \right]
$$
where each gradient term is given by $\delta^{(i)} = \nabla_\theta\mathcal{L} \left( f^\theta \left( x^{(i)} \right), y^{(i)} \right)$. Our goal is to bound this descent direction for the case when (up to) $n$ points are removed or added to the training data, for any $\theta \in [\theta_L, \theta_U]$. We begin by bounding the descent direction for a fixed, scalar $\theta$, then generalize to all $\theta \in [\theta_L, \theta_U]$ and to the multi-dimensional case (i.e., multiple parameters). We present only the upper bounds here; analogous results apply for lower bounds.

\textbf{Bounding the descent direction for a fixed, scalar $\theta$.} Consider the effect of removing up to $n$ data points from batch $\mathcal{B}$. Without loss of generality, assume the gradient terms are sorted in descending order, i.e., $\delta^{(1)} \geq \delta^{(2)} \geq \dots \geq \delta^{(b)}$. Then, the average clipped gradient over all points can be bounded above by the average over the largest $b - n$ terms:
$$
\Delta \theta = \frac{1}{b} \sum_{i=1}^b \operatorname{Clip}_\kappa \left[ \delta^{(i)} \right] \leq \frac{1}{b - n} \sum_{i=1}^{b - n} \operatorname{Clip}_\kappa \left[ \delta^{(i)} \right]
$$
This bound corresponds to removing the $n$ points with the smallest gradients.

Next, consider adding $n$ arbitrary points to the training batch. Since each added point contributes at most $\kappa$ due to clipping, the descent direction with up to $n$ removals and $n$ additions is bounded by
$$
\frac{1}{b} \sum_{i=1}^b \operatorname{Clip}_\kappa \left[ \delta^{(i)} \right] \leq \frac{1}{b - n} \sum_{i=1}^{b - n} \operatorname{Clip}_\kappa \left[ \delta^{(i)} \right] \leq \frac{1}{b} \left( n \kappa + \sum_{i=1}^{b} \operatorname{Clip}_\kappa \left[ \delta^{(i)} \right] \right)
$$
where the bound now accounts for replacing the $n$ smallest gradient terms with the maximum possible value of $\kappa$ from the added samples.

\textbf{Bounding the effect of a variable parameter interval.} We extend this bound to any $\theta \in [\theta_L, \theta_U]$. Assume the existence of upper bounds $\delta^{(i)}_U$ on the clipped gradients for each data point over the interval, such that
$$
\delta^{(i)}_U \geq \operatorname{Clip}_\kappa \left[ \nabla_{\theta'} \mathcal{L} \left( f^{\theta'}(x^{(i)}), y^{(i)} \right) \right] \quad \forall \, \theta' \in [\theta_L, \theta_U].
$$
Then, using these upper bounds, we further bound $\Delta \theta$ as
$$
\Delta \theta \leq \frac{1}{b} \left( n \kappa + \sum_{i=1}^b \operatorname{Clip}_\kappa \left[ \delta^{(i)}_U \right] \right)
$$
where, as before, we assume $\delta^{(i)}_U$ are indexed in descending order.

\textbf{Extending to the multi-dimensional case.} To generalize to the multi-dimensional case, we apply the above bound component-wise. Since gradients are not necessarily ordered for each parameter component, we introduce the $\operatorname{SEMax}_n$ operator, which selects and sums the largest $n$ terms at each index. This yields the following bound on the descent direction:
$$
\Delta \theta \leq \frac{1}{b} \left( \underset{b-n}{\operatorname{SEMax}} \left\{ \delta^{(i)}_U \right\}_{i=1}^b + n \kappa \mathbf{1}_d \right)
$$
which holds for any $\theta \in [\theta_L, \theta_U]$ and up to $n$ removed and replaced points.
\qed

We have established the upper bound on the descent direction. The corresponding lower bound can be derived by reversing the inequalities and substituting $\operatorname{SEMax}$ with the analagous minimization operator, $\operatorname{SEMin}$.

\subsection{Proof of Theorem~\ref{thm:aggregation_poisoning} (Sound Aggregation for Bounded Perturbation)}

The nominal descent direction for a parameter $\theta$ is the averaged gradient over a training batch $\mathcal{B}$, defined as
$$
\Delta \theta = \frac{1}{b} \sum_{i=1}^b \delta^{(i)}
$$
where each gradient term is given by $\delta^{(i)} = \nabla_\theta \mathcal{L} \left( f^\theta \left( x^{(i)} \right), y^{(i)} \right)$.
Our goal is to upper bound this descent direction when up to $n$ points from $\mathcal{B}$ are poisoned by up to $\epsilon$ in the feature space and $\nu$ in label space. 
The bound is additionally computed with respect to any $\theta \in [\theta_L, \theta_U]$. We again begin by bounding the descent direction for a fixed $\theta$, then generalize to all $\theta \in [\theta_L, \theta_U]$. We present only the upper bounds here, though corresponding results for the lower bound can be shown by reversing the inequalities and replacing $\operatorname{SEMax}$ with $\operatorname{SEMin}$.

\textbf{Bounding the descent direction for a fixed $\theta$.} Consider the effect of poisoning either the features or the labels of a data point.
For a given data point, an adversary may choose to poison its features, its labels, or both. In total, at most $n$ points may be influenced by the poisoning adversary.
We assume that $n \leq b$, otherwise take at most $\min(n, b)$ points to be poisoned.

Assume that we have access to sound gradient upper bounds 
$$
\delta^{(i)} \leq \tilde{\delta}^{(i)}_U \quad \forall         \delta^{(i)} \in 
\left\{\nabla_{\theta'}  \mathcal{L} \left( f^{\theta'}\left(\tilde{x}\right), \tilde{y}\right)
\mid
\|x^{(i)} - \tilde{x} \|_p \leq \epsilon,
\|y^{(i)} - \tilde{y} \|_q \leq \nu
\right\}.
$$
where the inequalities are interpreted element-wise. Here, $\tilde{\delta}^{(i)}_U$ corresponds to an upper bound on the maximum possible gradient achievable at the data-point $(x^{(i)}, y^{(i)})$ through poisoning.

The adversary's maximum possible impact on the descent direction at any point $i$ is given by $\tilde{\delta}^{(i)}_U - \delta^{(i)}$. To maximize an upper bound on $\Delta\theta$, we consider the $n$ points with the largest possible adversarial contributions. Therefore, we obtain
$$
\Delta \theta = \frac{1}{b} \sum_{i=1}^b \delta^{(i)} \leq \frac{1}{b}\left(\underset{n}{\operatorname{SEMax}}
\left\{
\tilde{\delta}^{(i)}_U - {{\delta}^{(i)}}\right\}_{i=1}^b + \sum\limits_{i=1}^b {{\delta}^{(i)}}\right),
$$
where the $\operatorname{SEMax}$ operation corresponds to taking the sum of the largest $n$ elements of its argument at each element. This bound captures the maximum increase in $\Delta\theta$ that an adversary can induce by poisoning up to $n$ data points.

\textbf{Bounding the effect of a variable parameter interval.} Now, we wish to compute a bound on $\Delta\theta$ for any $\theta \in [\theta_L, \theta_U]$. To achieve this, we extend our previous gradient bounds to account for the interval over our parameters. Specifically, we define upper bounds on the nominal and adversarially perturbed gradients that hold across the entire parameter interval:
\begin{align*}
{\delta} \leq {\delta}^{(i)}_U \quad \forall        
&{\delta} \in 
\left\{\nabla_{\theta'}  \mathcal{L} \left( f^{\theta'}(x^{(i)}), y^{(i)}\right)
\mid
\theta' \in [\theta^L, \theta^U]
\right\},\\
\tilde{\delta} \leq \tilde{\delta}^{(i)}_U \quad \forall         &\tilde{\delta} \in 
\left\{\nabla_{\theta'}  \mathcal{L} \left( f^{\theta'}\left(\tilde{x}\right), \tilde{y}\right)
\mid
\theta' \in [\theta^L, \theta^U],
\|x^{(i)} - \tilde{x} \|_p \leq \epsilon,
\|y^{(i)} - \tilde{y} \|_q \leq \nu
\right\}.
\end{align*}
Thus, the descent direction is upper bounded by
$$\Delta\theta \leq \Delta \theta^U = \frac{1}{b}\left(\underset{n}{\operatorname{SEMax}}
\left\{
\tilde{\delta}^{(i)}_U - {{\delta}^{(i)}_U}\right\}_{i=1}^b + \sum\limits_{i=1}^b {{\delta}^{(i)}_U}
\right)$$
for all $\theta \in [\theta_L, \theta_U]$, where the appropriate bounds with respect to the parameter interval have been substituted in.

%% file: main.bbl
\begin{thebibliography}{76}
\providecommand{\natexlab}[1]{#1}
\providecommand{\url}[1]{\texttt{#1}}
\expandafter\ifx\csname urlstyle\endcsname\relax
  \providecommand{\doi}[1]{doi: #1}\else
  \providecommand{\doi}{doi: \begingroup \urlstyle{rm}\Url}\fi

\bibitem[Abadi et~al.(2016)Abadi, Chu, Goodfellow, McMahan, Mironov, Talwar, and Zhang]{abadi2016deep}
Martin Abadi, Andy Chu, Ian Goodfellow, H~Brendan McMahan, Ilya Mironov, Kunal Talwar, and Li~Zhang.
\newblock Deep learning with differential privacy.
\newblock In \emph{Proceedings of the 2016 ACM SIGSAC conference on computer and communications security}, pages 308--318, 2016.

\bibitem[Adams et~al.(2023)Adams, Patane, Lahijanian, and Laurenti]{adams2023bnn}
Steven Adams, Andrea Patane, Morteza Lahijanian, and Luca Laurenti.
\newblock Bnn-dp: robustness certification of bayesian neural networks via dynamic programming.
\newblock In \emph{International Conference on Machine Learning}, pages 133--151. PMLR, 2023.

\bibitem[Althoff et~al.(2021)Althoff, Frehse, and Girard]{althoff2021set}
Matthias Althoff, Goran Frehse, and Antoine Girard.
\newblock Set propagation techniques for reachability analysis.
\newblock \emph{Annual Review of Control, Robotics, and Autonomous Systems}, 4\penalty0 (1):\penalty0 369--395, 2021.

\bibitem[Anderson et~al.(2020)Anderson, Huchette, Ma, Tjandraatmadja, and Vielma]{anderson2020strong}
Ross Anderson, Joey Huchette, Will Ma, Christian Tjandraatmadja, and Juan~Pablo Vielma.
\newblock Strong mixed-integer programming formulations for trained neural networks.
\newblock \emph{Mathematical Programming}, 183\penalty0 (1):\penalty0 3--39, 2020.

\bibitem[Basu et~al.(2020)Basu, Pope, and Feizi]{basu2020influence}
Samyadeep Basu, Philip Pope, and Soheil Feizi.
\newblock Influence functions in deep learning are fragile.
\newblock \emph{arXiv preprint arXiv:2006.14651}, 2020.

\bibitem[Biggio and Roli(2018)]{biggio2018wild}
Battista Biggio and Fabio Roli.
\newblock Wild patterns: Ten years after the rise of adversarial machine learning.
\newblock In \emph{Proceedings of the 2018 ACM SIGSAC Conference on Computer and Communications Security}, pages 2154--2156, 2018.

\bibitem[Biggio et~al.(2014)Biggio, Rieck, Ariu, Wressnegger, Corona, Giacinto, and Roli]{biggio2014poisoning}
Battista Biggio, Konrad Rieck, Davide Ariu, Christian Wressnegger, Igino Corona, Giorgio Giacinto, and Fabio Roli.
\newblock Poisoning behavioral malware clustering.
\newblock In \emph{Proceedings of the 2014 workshop on artificial intelligent and security workshop}, pages 27--36, 2014.

\bibitem[Bojarski et~al.(2016)Bojarski, Del~Testa, Dworakowski, Firner, Flepp, Goyal, Jackel, Monfort, Muller, Zhang, et~al.]{bojarski2016end}
Mariusz Bojarski, Davide Del~Testa, Daniel Dworakowski, Bernhard Firner, Beat Flepp, Prasoon Goyal, Lawrence~D Jackel, Mathew Monfort, Urs Muller, Jiakai Zhang, et~al.
\newblock End to end learning for self-driving cars.
\newblock \emph{arXiv preprint arXiv:1604.07316}, 2016.

\bibitem[Bommasani et~al.(2021)Bommasani, Hudson, Adeli, Altman, Arora, von Arx, Bernstein, Bohg, Bosselut, Brunskill, et~al.]{bommasani2021opportunities}
Rishi Bommasani, Drew~A Hudson, Ehsan Adeli, Russ Altman, Simran Arora, Sydney von Arx, Michael~S Bernstein, Jeannette Bohg, Antoine Bosselut, Emma Brunskill, et~al.
\newblock On the opportunities and risks of foundation models.
\newblock \emph{arXiv preprint arXiv:2108.07258}, 2021.

\bibitem[Bourtoule et~al.(2021)Bourtoule, Chandrasekaran, Choquette-Choo, Jia, Travers, Zhang, Lie, and Papernot]{bourtoule2021machine}
Lucas Bourtoule, Varun Chandrasekaran, Christopher~A Choquette-Choo, Hengrui Jia, Adelin Travers, Baiwu Zhang, David Lie, and Nicolas Papernot.
\newblock Machine unlearning.
\newblock In \emph{2021 IEEE symposium on security and privacy (SP)}, pages 141--159. IEEE, 2021.

\bibitem[Bunel et~al.(2018)Bunel, Turkaslan, Torr, Kohli, and Mudigonda]{bunel2018unified}
Rudy~R Bunel, Ilker Turkaslan, Philip Torr, Pushmeet Kohli, and Pawan~K Mudigonda.
\newblock A unified view of piecewise linear neural network verification.
\newblock \emph{Advances in Neural Information Processing Systems}, 31, 2018.

\bibitem[Cao and Yang(2015)]{cao2015towards}
Yinzhi Cao and Junfeng Yang.
\newblock Towards making systems forget with machine unlearning.
\newblock In \emph{2015 IEEE symposium on security and privacy}, pages 463--480. IEEE, 2015.

\bibitem[Carlini et~al.(2022)Carlini, Chien, Nasr, Song, Terzis, and Tramer]{carlini2022membership}
Nicholas Carlini, Steve Chien, Milad Nasr, Shuang Song, Andreas Terzis, and Florian Tramer.
\newblock Membership inference attacks from first principles.
\newblock In \emph{2022 IEEE Symposium on Security and Privacy (SP)}, pages 1897--1914. IEEE, 2022.

\bibitem[Carlini et~al.(2023)Carlini, Jagielski, Choquette-Choo, Paleka, Pearce, Anderson, Terzis, Thomas, and Tram{\`e}r]{carlini2023poisoning}
Nicholas Carlini, Matthew Jagielski, Christopher~A Choquette-Choo, Daniel Paleka, Will Pearce, Hyrum Anderson, Andreas Terzis, Kurt Thomas, and Florian Tram{\`e}r.
\newblock Poisoning web-scale training datasets is practical.
\newblock \emph{arXiv preprint arXiv:2302.10149}, 2023.

\bibitem[Chadha et~al.(2024)Chadha, Jagielski, Papernot, Choquette-Choo, and Nasr]{chadha2024auditing}
Karan Chadha, Matthew Jagielski, Nicolas Papernot, Christopher Choquette-Choo, and Milad Nasr.
\newblock Auditing private prediction.
\newblock \emph{arXiv preprint arXiv:2402.09403}, 2024.

\bibitem[Chen et~al.(2022)Chen, Li, Li, Yan, and Wu]{chen2022collective}
Ruoxin Chen, Zenan Li, Jie Li, Junchi Yan, and Chentao Wu.
\newblock On collective robustness of bagging against data poisoning.
\newblock In \emph{International Conference on Machine Learning}, pages 3299--3319. PMLR, 2022.

\bibitem[Chen et~al.(2017)Chen, Liu, Li, Lu, and Song]{chen2017targeted}
Xinyun Chen, Chang Liu, Bo~Li, Kimberly Lu, and Dawn Song.
\newblock Targeted backdoor attacks on deep learning systems using data poisoning.
\newblock \emph{arXiv preprint arXiv:1712.05526}, 2017.

\bibitem[Cummings et~al.(2021)Cummings, Kaptchuk, and Redmiles]{cummings2021need}
Rachel Cummings, Gabriel Kaptchuk, and Elissa~M Redmiles.
\newblock " i need a better description": An investigation into user expectations for differential privacy.
\newblock In \emph{Proceedings of the 2021 ACM SIGSAC Conference on Computer and Communications Security}, pages 3037--3052, 2021.

\bibitem[Dwork et~al.(2014)Dwork, Roth, et~al.]{dwork2014algorithmic}
Cynthia Dwork, Aaron Roth, et~al.
\newblock The algorithmic foundations of differential privacy.
\newblock \emph{Foundations and Trends{\textregistered} in Theoretical Computer Science}, 9\penalty0 (3--4):\penalty0 211--407, 2014.

\bibitem[Ehlers(2017)]{ehlers_formal_2017}
Ruediger Ehlers.
\newblock Formal {Verification} of {Piece}-{Wise} {Linear} {Feed}-{Forward} {Neural} {Networks}, August 2017.
\newblock URL \url{http://arxiv.org/abs/1705.01320}.
\newblock arXiv:1705.01320 [cs].

\bibitem[Fioretto et~al.(2022)Fioretto, Tran, Van~Hentenryck, and Zhu]{fioretto2022differential}
Ferdinando Fioretto, Cuong Tran, Pascal Van~Hentenryck, and Keyu Zhu.
\newblock Differential privacy and fairness in decisions and learning tasks: A survey.
\newblock \emph{arXiv preprint arXiv:2202.08187}, 2022.

\bibitem[Fischetti and Jo(2018)]{fischetti2018deep}
Matteo Fischetti and Jason Jo.
\newblock Deep neural networks and mixed integer linear optimization.
\newblock \emph{Constraints}, 23\penalty0 (3):\penalty0 296--309, 2018.

\bibitem[Gowal et~al.(2018)Gowal, Dvijotham, Stanforth, Bunel, Qin, Uesato, Arandjelovic, Mann, and Kohli]{gowal2018effectiveness}
Sven Gowal, Krishnamurthy Dvijotham, Robert Stanforth, Rudy Bunel, Chongli Qin, Jonathan Uesato, Relja Arandjelovic, Timothy Mann, and Pushmeet Kohli.
\newblock On the effectiveness of interval bound propagation for training verifiably robust models.
\newblock \emph{arXiv preprint arXiv:1810.12715}, 2018.

\bibitem[Gu et~al.(2017)Gu, Dolan-Gavitt, and Garg]{gu2017badnets}
Tianyu Gu, Brendan Dolan-Gavitt, and Siddharth Garg.
\newblock Badnets: Identifying vulnerabilities in the machine learning model supply chain.
\newblock \emph{arXiv preprint arXiv:1708.06733}, 2017.

\bibitem[Guo et~al.(2019)Guo, Goldstein, Hannun, and Van Der~Maaten]{guo2019certified}
Chuan Guo, Tom Goldstein, Awni Hannun, and Laurens Van Der~Maaten.
\newblock Certified data removal from machine learning models.
\newblock \emph{arXiv preprint arXiv:1911.03030}, 2019.

\bibitem[Han et~al.(2022)Han, Xu, Zhou, Yang, Li, and Zhang]{han2022physical}
Xingshuo Han, Guowen Xu, Yuan Zhou, Xuehuan Yang, Jiwei Li, and Tianwei Zhang.
\newblock Physical backdoor attacks to lane detection systems in autonomous driving.
\newblock In \emph{Proceedings of the 30th ACM International Conference on Multimedia}, pages 2957--2968, 2022.

\bibitem[Hebrail and Berard(2012)]{misc_individual_household_electric_power_consumption_235}
Georges Hebrail and Alice Berard.
\newblock {Individual Household Electric Power Consumption}.
\newblock UCI Machine Learning Repository, 2012.
\newblock {DOI}: https://doi.org/10.24432/C58K54.

\bibitem[Hong et~al.(2020)Hong, Chandrasekaran, Kaya, Dumitra{\c{s}}, and Papernot]{hong2020effectiveness}
Sanghyun Hong, Varun Chandrasekaran, Yi{\u{g}}itcan Kaya, Tudor Dumitra{\c{s}}, and Nicolas Papernot.
\newblock On the effectiveness of mitigating data poisoning attacks with gradient shaping.
\newblock \emph{arXiv preprint arXiv:2002.11497}, 2020.

\bibitem[Huchette et~al.(2023)Huchette, Mu{\~n}oz, Serra, and Tsay]{huchette2023deep}
Joey Huchette, Gonzalo Mu{\~n}oz, Thiago Serra, and Calvin Tsay.
\newblock When deep learning meets polyhedral theory: A survey.
\newblock \emph{arXiv preprint arXiv:2305.00241}, 2023.

\bibitem[Katz et~al.(2017)Katz, Barrett, Dill, Julian, and Kochenderfer]{katz2017reluplex}
Guy Katz, Clark Barrett, David~L Dill, Kyle Julian, and Mykel~J Kochenderfer.
\newblock Reluplex: An efficient smt solver for verifying deep neural networks.
\newblock In \emph{Computer Aided Verification: 29th International Conference, CAV 2017, Heidelberg, Germany, July 24-28, 2017, Proceedings, Part I 30}, pages 97--117. Springer, 2017.

\bibitem[Kingma and Ba(2014)]{kingma2014adam}
Diederik~P Kingma and Jimmy Ba.
\newblock Adam: A method for stochastic optimization.
\newblock \emph{arXiv preprint arXiv:1412.6980}, 2014.

\bibitem[Koh and Liang(2017)]{koh2017understanding}
Pang~Wei Koh and Percy Liang.
\newblock Understanding black-box predictions via influence functions.
\newblock In \emph{International conference on machine learning}, pages 1885--1894. PMLR, 2017.

\bibitem[K{\"o}nig et~al.(2024)K{\"o}nig, Bosman, Hoos, and van Rijn]{konig2024critically}
Matthias K{\"o}nig, Annelot~W Bosman, Holger~H Hoos, and Jan~N van Rijn.
\newblock Critically assessing the state of the art in neural network verification.
\newblock \emph{Journal of Machine Learning Research}, 25\penalty0 (12):\penalty0 1--53, 2024.

\bibitem[Levine and Feizi(2020)]{levine2020deep}
Alexander Levine and Soheil Feizi.
\newblock Deep partition aggregation: Provable defense against general poisoning attacks.
\newblock \emph{arXiv preprint arXiv:2006.14768}, 2020.

\bibitem[Li et~al.(2020)Li, Cheng, Wang, Liu, and Chen]{li2020learning}
Suyi Li, Yong Cheng, Wei Wang, Yang Liu, and Tianjian Chen.
\newblock Learning to detect malicious clients for robust federated learning.
\newblock \emph{arXiv preprint arXiv:2002.00211}, 2020.

\bibitem[Ligett et~al.(2017)Ligett, Neel, Roth, Waggoner, and Wu]{ligett2017accuracy}
Katrina Ligett, Seth Neel, Aaron Roth, Bo~Waggoner, and Steven~Z Wu.
\newblock Accuracy first: Selecting a differential privacy level for accuracy constrained erm.
\newblock \emph{Advances in Neural Information Processing Systems}, 30, 2017.

\bibitem[Liu et~al.(2019)Liu, Arnon, Lazarus, Strong, Barrett, and Kochenderfer]{liu_algorithms_2019}
Changliu Liu, Tomer Arnon, Christopher Lazarus, Christopher Strong, Clark Barrett, and Mykel~J. Kochenderfer.
\newblock Algorithms for {Verifying} {Deep} {Neural} {Networks}, March 2019.
\newblock URL \url{https://arxiv.org/abs/1903.06758v2}.

\bibitem[Liu et~al.(2022)Liu, Kong, and Oh]{liu2022differential}
Xiyang Liu, Weihao Kong, and Sewoong Oh.
\newblock Differential privacy and robust statistics in high dimensions.
\newblock In \emph{Conference on Learning Theory}, pages 1167--1246. PMLR, 2022.

\bibitem[Lorenz et~al.(2024{\natexlab{a}})Lorenz, Kwiatkowska, and Fritz]{lorenz2024bicert}
Tobias Lorenz, Marta Kwiatkowska, and Mario Fritz.
\newblock Bicert: A bilinear mixed integer programming formulation for precise certified bounds against data poisoning attacks.
\newblock \emph{arXiv preprint arXiv:2412.10186}, 2024{\natexlab{a}}.

\bibitem[Lorenz et~al.(2024{\natexlab{b}})Lorenz, Kwiatkowska, and Fritz]{lorenz2024fullcert}
Tobias Lorenz, Marta Kwiatkowska, and Mario Fritz.
\newblock Fullcert: Deterministic end-to-end certification for training and inference of neural networks.
\newblock In \emph{DAGM German Conference on Pattern Recognition}, pages 71--85. Springer, 2024{\natexlab{b}}.

\bibitem[Maas et~al.(2011)Maas, Daly, Pham, Huang, Ng, and Potts]{imdb}
Andrew~L. Maas, Raymond~E. Daly, Peter~T. Pham, Dan Huang, Andrew~Y. Ng, and Christopher Potts.
\newblock Learning word vectors for sentiment analysis.
\newblock In \emph{Proceedings of the 49th Annual Meeting of the Association for Computational Linguistics: Human Language Technologies}, pages 142--150, Portland, Oregon, USA, June 2011. Association for Computational Linguistics.
\newblock URL \url{http://www.aclweb.org/anthology/P11-1015}.

\bibitem[McCormick(1976)]{mccormick1976computability}
Garth~P McCormick.
\newblock Computability of global solutions to factorable nonconvex programs: Part i—convex underestimating problems.
\newblock \emph{Mathematical programming}, 10\penalty0 (1):\penalty0 147--175, 1976.

\bibitem[Mu{\~n}oz-Gonz{\'a}lez et~al.(2017)Mu{\~n}oz-Gonz{\'a}lez, Biggio, Demontis, Paudice, Wongrassamee, Lupu, and Roli]{munoz2017towards}
Luis Mu{\~n}oz-Gonz{\'a}lez, Battista Biggio, Ambra Demontis, Andrea Paudice, Vasin Wongrassamee, Emil~C Lupu, and Fabio Roli.
\newblock Towards poisoning of deep learning algorithms with back-gradient optimization.
\newblock In \emph{Proceedings of the 10th ACM workshop on artificial intelligence and security}, pages 27--38, 2017.

\bibitem[Neel et~al.(2021)Neel, Roth, and Sharifi-Malvajerdi]{neel2021descent}
Seth Neel, Aaron Roth, and Saeed Sharifi-Malvajerdi.
\newblock Descent-to-delete: Gradient-based methods for machine unlearning.
\newblock In \emph{Algorithmic Learning Theory}, pages 931--962. PMLR, 2021.

\bibitem[Newsome et~al.(2006)Newsome, Karp, and Song]{newsome2006paragraph}
James Newsome, Brad Karp, and Dawn Song.
\newblock Paragraph: Thwarting signature learning by training maliciously.
\newblock In \emph{Recent Advances in Intrusion Detection: 9th International Symposium, RAID 2006 Hamburg, Germany, September 20-22, 2006 Proceedings 9}, pages 81--105. Springer, 2006.

\bibitem[Nguyen(2012)]{nguyen2012efficient}
Hong~Diep Nguyen.
\newblock Efficient implementation of interval matrix multiplication.
\newblock In \emph{Applied Parallel and Scientific Computing: 10th International Conference, PARA 2010, Reykjav{\'\i}k, Iceland, June 6-9, 2010, Revised Selected Papers, Part II 10}, pages 179--188. Springer, 2012.

\bibitem[Nissim et~al.(2007)Nissim, Raskhodnikova, and Smith]{nissim2007smooth}
Kobbi Nissim, Sofya Raskhodnikova, and Adam Smith.
\newblock Smooth sensitivity and sampling in private data analysis.
\newblock In \emph{Proceedings of the thirty-ninth annual ACM symposium on Theory of computing}, pages 75--84, 2007.

\bibitem[Papernot et~al.(2016{\natexlab{a}})Papernot, Abadi, Erlingsson, Goodfellow, and Talwar]{papernot2016semi}
Nicolas Papernot, Mart{\'\i}n Abadi, Ulfar Erlingsson, Ian Goodfellow, and Kunal Talwar.
\newblock Semi-supervised knowledge transfer for deep learning from private training data.
\newblock \emph{arXiv preprint arXiv:1610.05755}, 2016{\natexlab{a}}.

\bibitem[Papernot et~al.(2016{\natexlab{b}})Papernot, McDaniel, Jha, Fredrikson, Celik, and Swami]{papernot2016limitations}
Nicolas Papernot, Patrick McDaniel, Somesh Jha, Matt Fredrikson, Z~Berkay Celik, and Ananthram Swami.
\newblock The limitations of deep learning in adversarial settings.
\newblock In \emph{2016 IEEE European symposium on security and privacy (EuroS\&P)}, pages 372--387. IEEE, 2016{\natexlab{b}}.

\bibitem[Radford et~al.(2019)Radford, Wu, Child, Luan, Amodei, Sutskever, et~al.]{radford2019language}
Alec Radford, Jeffrey Wu, Rewon Child, David Luan, Dario Amodei, Ilya Sutskever, et~al.
\newblock Language models are unsupervised multitask learners.
\newblock \emph{OpenAI blog}, 1\penalty0 (8):\penalty0 9, 2019.

\bibitem[Rezaei et~al.(2023)Rezaei, Banihashem, Chegini, and Feizi]{rezaei2023run}
Keivan Rezaei, Kiarash Banihashem, Atoosa Chegini, and Soheil Feizi.
\newblock Run-off election: Improved provable defense against data poisoning attacks.
\newblock In \emph{International Conference on Machine Learning}, pages 29030--29050. PMLR, 2023.

\bibitem[Rosenfeld et~al.(2020)Rosenfeld, Winston, Ravikumar, and Kolter]{rosenfeld2020certified}
Elan Rosenfeld, Ezra Winston, Pradeep Ravikumar, and Zico Kolter.
\newblock Certified robustness to label-flipping attacks via randomized smoothing.
\newblock In \emph{International Conference on Machine Learning}, pages 8230--8241. PMLR, 2020.

\bibitem[Rump(1999)]{rump1999fast}
Siegfried~M Rump.
\newblock Fast and parallel interval arithmetic.
\newblock \emph{BIT Numerical Mathematics}, 39:\penalty0 534--554, 1999.

\bibitem[Singh et~al.(2019)Singh, Gehr, P\"{u}schel, and Vechev]{singh2019}
Gagandeep Singh, Timon Gehr, Markus P\"{u}schel, and Martin Vechev.
\newblock An abstract domain for certifying neural networks.
\newblock \emph{Proc. ACM Program. Lang.}, 3\penalty0 (POPL), jan 2019.
\newblock \doi{10.1145/3290354}.

\bibitem[Sosnin and Tsay(2024)]{sosnin2024scaling}
Philip Sosnin and Calvin Tsay.
\newblock Scaling mixed-integer programming for certification of neural network controllers using bounds tightening.
\newblock \emph{arXiv preprint arXiv:2403.17874}, 2024.

\bibitem[Sosnin et~al.(2024)Sosnin, M{\"u}ller, Baader, Tsay, and Wicker]{sosnin2024certified}
Philip Sosnin, Mark~N M{\"u}ller, Maximilian Baader, Calvin Tsay, and Matthew Wicker.
\newblock Certified robustness to data poisoning in gradient-based training.
\newblock \emph{arXiv preprint arXiv:2406.05670}, 2024.

\bibitem[Steinhardt et~al.(2017)Steinhardt, Koh, and Liang]{steinhardt2017certified}
Jacob Steinhardt, Pang Wei~W Koh, and Percy~S Liang.
\newblock Certified defenses for data poisoning attacks.
\newblock \emph{Advances in neural information processing systems}, 30, 2017.

\bibitem[Szegedy et~al.(2013)Szegedy, Zaremba, Sutskever, Bruna, Erhan, Goodfellow, and Fergus]{szegedy2013intriguing}
Christian Szegedy, Wojciech Zaremba, Ilya Sutskever, Joan Bruna, Dumitru Erhan, Ian Goodfellow, and Rob Fergus.
\newblock Intriguing properties of neural networks.
\newblock \emph{arXiv preprint arXiv:1312.6199}, 2013.

\bibitem[Thudi et~al.(2022)Thudi, Jia, Shumailov, and Papernot]{thudi2022necessity}
Anvith Thudi, Hengrui Jia, Ilia Shumailov, and Nicolas Papernot.
\newblock On the necessity of auditable algorithmic definitions for machine unlearning.
\newblock In \emph{31st USENIX security symposium (USENIX Security 22)}, pages 4007--4022, 2022.

\bibitem[Tian et~al.(2022)Tian, Cui, Liang, and Yu]{tian2022comprehensive}
Zhiyi Tian, Lei Cui, Jie Liang, and Shui Yu.
\newblock A comprehensive survey on poisoning attacks and countermeasures in machine learning.
\newblock \emph{ACM Computing Surveys}, 55\penalty0 (8):\penalty0 1--35, 2022.

\bibitem[Tsay et~al.(2021)Tsay, Kronqvist, Thebelt, and Misener]{tsay2021partition}
Calvin Tsay, Jan Kronqvist, Alexander Thebelt, and Ruth Misener.
\newblock Partition-based formulations for mixed-integer optimization of trained relu neural networks.
\newblock \emph{Advances in neural information processing systems}, 34:\penalty0 3068--3080, 2021.

\bibitem[van~der Maaten and Hannun(2020)]{van2020trade}
Laurens van~der Maaten and Awni Hannun.
\newblock The trade-offs of private prediction.
\newblock \emph{arXiv preprint arXiv:2007.05089}, 2020.

\bibitem[Wang et~al.(2022)Wang, Levine, and Feizi]{wang2022improved}
Wenxiao Wang, Alexander~J Levine, and Soheil Feizi.
\newblock Improved certified defenses against data poisoning with (deterministic) finite aggregation.
\newblock In \emph{International Conference on Machine Learning}, pages 22769--22783. PMLR, 2022.

\bibitem[Wei et~al.(2023)Wei, Wu, Wu, Chen, Barrett, and Farchi]{wei2023convex}
Dennis Wei, Haoze Wu, Min Wu, Pin-Yu Chen, Clark Barrett, and Eitan Farchi.
\newblock Convex bounds on the softmax function with applications to robustness verification, 2023.

\bibitem[Wicker et~al.(2020)Wicker, Laurenti, Patane, and Kwiatkowska]{wicker2020probabilistic}
Matthew Wicker, Luca Laurenti, Andrea Patane, and Marta Kwiatkowska.
\newblock Probabilistic safety for bayesian neural networks.
\newblock In \emph{Conference on uncertainty in artificial intelligence}, pages 1198--1207. PMLR, 2020.

\bibitem[Wicker et~al.(2022)Wicker, Heo, Costabello, and Weller]{wicker2022robust}
Matthew Wicker, Juyeon Heo, Luca Costabello, and Adrian Weller.
\newblock Robust explanation constraints for neural networks.
\newblock \emph{arXiv preprint arXiv:2212.08507}, 2022.

\bibitem[Wicker et~al.(2023)Wicker, Patane, Laurenti, and Kwiatkowska]{wicker2023adversarial}
Matthew Wicker, Andrea Patane, Luca Laurenti, and Marta Kwiatkowska.
\newblock Adversarial robustness certification for bayesian neural networks.
\newblock \emph{arXiv preprint arXiv:2306.13614}, 2023.

\bibitem[Wicker et~al.(2024)Wicker, Sosnin, Shilov, Janik, M{\"u}ller, de~Montjoye, Weller, and Tsay]{wicker2024certification}
Matthew Wicker, Philip Sosnin, Igor Shilov, Adrianna Janik, Mark~N M{\"u}ller, Yves-Alexandre de~Montjoye, Adrian Weller, and Calvin Tsay.
\newblock Certification for differentially private prediction in gradient-based training.
\newblock \emph{arXiv preprint arXiv:2406.13433}, 2024.

\bibitem[Xiang et~al.(2018)Xiang, Tran, and Johnson]{xiang2018output}
Weiming Xiang, Hoang-Dung Tran, and Taylor~T Johnson.
\newblock Output reachable set estimation and verification for multilayer neural networks.
\newblock \emph{IEEE transactions on neural networks and learning systems}, 29\penalty0 (11):\penalty0 5777--5783, 2018.

\bibitem[Xie et~al.(2022)Xie, Long, Chen, and Li]{xie2022uncovering}
Chulin Xie, Yunhui Long, Pin-Yu Chen, and Bo~Li.
\newblock Uncovering the connection between differential privacy and certified robustness of federated learning against poisoning attacks.
\newblock \emph{arXiv preprint arXiv:2209.04030}, 2022.

\bibitem[Yang et~al.(2017)Yang, Gong, and Cai]{yang2017fake}
Guolei Yang, Neil~Zhenqiang Gong, and Ying Cai.
\newblock Fake co-visitation injection attacks to recommender systems.
\newblock In \emph{NDSS}, 2017.

\bibitem[Yang et~al.(2021)Yang, Shi, and Ni]{medmnist}
Jiancheng Yang, Rui Shi, and Bingbing Ni.
\newblock {MedMNIST} classification decathlon: A lightweight {AutoML} benchmark for medical image analysis.
\newblock In \emph{IEEE 18th International Symposium on Biomedical Imaging (ISBI)}, pages 191--195, 2021.

\bibitem[Yu et~al.(2022)Yu, Kamath, Kulkarni, Liu, Yin, and Zhang]{yu2022individual}
Da~Yu, Gautam Kamath, Janardhan Kulkarni, Tie-Yan Liu, Jian Yin, and Huishuai Zhang.
\newblock Individual privacy accounting for differentially private stochastic gradient descent.
\newblock \emph{arXiv preprint arXiv:2206.02617}, 2022.

\bibitem[Zhang et~al.(2018)Zhang, Weng, Chen, Hsieh, and Daniel]{zhang2018efficient}
Huan Zhang, Tsui-Wei Weng, Pin-Yu Chen, Cho-Jui Hsieh, and Luca Daniel.
\newblock Efficient neural network robustness certification with general activation functions.
\newblock \emph{Advances in neural information processing systems}, 31, 2018.

\bibitem[Zhang et~al.(2020)Zhang, Zhu, and Lessard]{zhang2020online}
Xuezhou Zhang, Xiaojin Zhu, and Laurent Lessard.
\newblock Online data poisoning attacks.
\newblock In \emph{Learning for Dynamics and Control}, pages 201--210. PMLR, 2020.

\bibitem[Zhu et~al.(2019)Zhu, Huang, Li, Taylor, Studer, and Goldstein]{zhu2019transferable}
Chen Zhu, W~Ronny Huang, Hengduo Li, Gavin Taylor, Christoph Studer, and Tom Goldstein.
\newblock Transferable clean-label poisoning attacks on deep neural nets.
\newblock In \emph{International conference on machine learning}, pages 7614--7623. PMLR, 2019.

\end{thebibliography}
